\documentclass[runningheads]{llncs}
\usepackage{cite}
\usepackage{graphicx}
\usepackage{xspace}
\usepackage{subfig}
\usepackage{standalone}
\usepackage{tikz}
\usetikzlibrary{positioning, shapes.geometric, shapes.arrows}
\usepackage{amsmath,amssymb}
\usepackage{float}
\usepackage{comment}

\newcommand{\model}{LSM\xspace} % 
\newcommand{\toy}{synthetic image translation\xspace}

\newcommand{\TOY}{Synthetic Image Translation\xspace}
\newcommand{\toyS}{SIT\xspace}
\newcommand{\sarcoS}{SSS\xspace}
\newcommand{\faceS}{300-W\xspace}
\newcommand{\px}{p(x)}
\newcommand{\py}{p(y)}
\newcommand{\pxy}{p(x,y)}
\newcommand{\ptx}{p(\tilde x)}
\newcommand{\pty}{p(\tilde y)}
\newcommand{\ptxy}{p(\tilde x,\tilde y)}
\newcommand{\dcy}{DC_{\tilde y}}
\newcommand{\dcx}{DC_{\tilde x}}
\newcommand{\Lxy}{L_{\tilde x \rightarrow \tilde y}}

\newcommand{\Lxyx}{L_{\tilde y \rightarrow \tilde x}}
\newcommand{\lossdist}{\mathcal{L}_{Dist_{\tilde y}}}
\newcommand{\lossconf}{\mathcal{L}_{Conf_{\tilde y}}}
\newcommand{\lossdistx}{\mathcal{L}_{Dist_{\tilde x}}}
\newcommand{\lossconfx}{\mathcal{L}_{Conf_{\tilde x}}}
\newcommand{\code}{https://gitlab.insa-rouen.fr/tmayet/lsm-latent-space-mapping}

\begin{document}
\title{Domain Translation via Latent Space Mapping}
\author{
Tsiry Mayet\inst{1} \and
Simon Bernard\inst{2} \and
Clement Chatelain\inst{1} \and
Romain Herault\inst{1}
}
\authorrunning{T. Mayet et al.}

\institute{
INSA Rouen Normandie, LITIS UR 4108, F-76000 Rouen, France \and
Univ Rouen Normandie, LITIS UR 4108, F-76000 Rouen, France
}

\maketitle
\begin{abstract}
In this paper, we investigate the problem of multi-domain translation: given an element $a$ of domain $A$, we would like to generate a corresponding $b$ sample in another domain $B$, and vice versa.
Acquiring supervision in multiple domains can be a tedious task, also we propose to learn this translation from one domain to another when supervision is available as a pair $(a,b)\sim A\times B$ and leveraging possible unpaired data when only $a\sim A$ or only $b\sim B$ is available.
We introduce a new unified framework called Latent Space Mapping (\model) that exploits the manifold assumption in order to learn, from each domain, a latent space. Unlike existing approaches, we propose to further regularize each latent space using available domains by learning each dependency between pairs of domains.
We evaluate our approach in three tasks performing i) synthetic dataset with image translation, ii) real-world task of semantic segmentation for medical images, and iii) real-world task of facial landmark detection. Source code is publicly available\footnote{\code}.

\end{abstract}

\section{Introduction}
% Presentation of the principle to have multi-domain, multi-views
In many machine learning tasks, different modalities can be modeled as different domains and different data as different views of the same reality. For example considering an autonomous vehicle context, the RGB camera, the depth map, and the segmentation map can be considered as three views of the same reality.
% No input or output
Often one domain can be considered as an input domain, e.g. a CT (computed tomography) scan of a patient, and other domains as an output domain, e.g. organ segmentation of the scan. Here we do not consider an input nor output domain, but rather we would like to learn to translate from any domain to another one. This definition masks the classical definition of an input and output domain, as every domain can be a possible input or output.
% DL approaches problems to succeed with a lot of data, but not always possible
While the rising amount of available data has brought great results in domain translation tasks in a fully supervised fashion, in some fields the quantity of available supervision is limited and hard to obtain, as in the medical field where a domain expert is required to create a hand-made segmentation \cite{IODA,every_annotation_count}, and this low supervision setting often reduces the performance of classical deep learning model.

Acquiring complete supervision to learn this translation across the different domains can be a tedious task, since for one view all the other corresponding views in the other domains are required. We now consider the setting of incomplete supervision, where for one view, the other view might or might not be available.

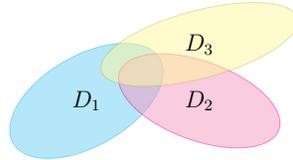
\begin{figure}
\centering
\begin{tikzpicture}
    \tikzstyle{every node}=[font=\large]

  \coordinate[] (a) at (0,0);
  \coordinate[right=20mm of a] (b);
  \coordinate[above=10mm of b] (c);
  
  \path[rotate=-60,draw=cyan,fill=cyan!50,opacity=0.5]  (a) ellipse (8mm and 15mm);
  \path[rotate=70,draw=magenta,fill=magenta!50,opacity=0.5]  (b) ellipse (7mm and 15mm);
  \path[rotate=-75,draw=yellow!75!black,fill=yellow!50,opacity=0.5] (c) ellipse (6mm and 18mm);
  
  \node at (a) {$D_1$};
  \node at (b) {$D_2$};
  \node at (c) {$D_3$};

\end{tikzpicture}
\caption{Set of three domains $D_1$, $D_2$ and $D_3$, each view can have or not have a corresponding view in the other domain, creating a situation of semi-supervised learning}
\label{fig:intro:3domains}
\end{figure}

% Presentation of the domain translation task
We investigate the task of multi-domain translation, i.e., given a set of $n$ domains $\{D_1, \ldots, D_n\}$, we would like to learn a mapping function $\forall i,j\;\;f_{i \rightarrow j}: D_i \rightarrow D_j$ allowing to translate a view from the domain $D_i$ into its view in $D_j$. An illustration with $n=3$ is given in Figure~\ref{fig:intro:3domains} where multiple possible configurations of the triple $(d_1,d_2,d_3)$ are possible, and each point does not necessarily have an associated view in the other two domains available (e.g. example of available data: $(d_1,d_2)$, $(d_2,d_3)$, or only $(d_1)$).

%add quick sentence to explain what other do
Many classical machine learning tasks fall into this domain translation framework, provided that one can reconstruct a view $d_i$ given another view $d_j$. For example, image captioning and machine translation can be viewed as domain translation problems. In Image captioning, one domain is the image and the other is its description, and for machine translation both domains are textual.
Previous approaches leverage useful assumptions and inductive bias, a common one in domain translation is the shared latent space assumption \cite{GRL,shared1corrnet,shared2alignflow,shared3,shared4} - associated views of different domains can be embedded as the same code in a latent space.

In this paper, we propose \model unified framework, which allows learning to translate one view from a domain to another domain in a semi-supervised setting when either all the views in each domain are available, or when an arbitrary number of views are available.
We do so by learning a latent space for each of the domains where each view can be embedded in a manifold.

Each manifold is further constrained by the mean of distance and adversarial losses allowing \model to learn a better representation by leveraging the intra-domain dependencies (dependencies within one domain, e.g. the value of one image pixel or segmentation mask regarding its surrounding pixel values) and inter-domain dependencies (the dependencies between multiple domains, e.g. the pixel of an image with regard to the associated pixel of the segmentation mask).

The rest of this paper is organized as follows, Section 2 reviews related work to domain translation, Section 3 presents how \model framework is structured and trained, and Section 4 describes how domain translation can be applied to three tasks (semantic segmentation, image to image translation, and facial landmark detection). The detailed settings of the training are provided in the Appendix.
\section{Related Works}
Domain translation is related to other tasks like semi-supervised learning (SSL), and domain adaptation (DA). While SSL and DA differ from domain translation, these three tasks leverage common assumptions and often use similar methods.
The goal of SSL is to learn from labeled and unlabeled samples, but in SSL an input domain and output domain are clearly identified.
In domain translation, a sample can be instantiated in $A$ but not in $B$, in $B$ but not in $A$, or in $A$ and $B$. None of the domains can be considered either as `input' nor `output'.
In DA, an input domain and an output domain are also often considered, but the input domain contains two sub-domains: a source domain and a target domain, in this setting close to SSL, the source domain contains labeled examples and the target domain is often left without supervision. The discrepancy between the two inputs domains prevents from just merging them together.
In this work, we leverage two useful SSL assumptions that we will present: the Manifold Assumption that we use to learn the intra-domain dependencies, and the Shared Latent Space Assumption that we relax in order to learn inter-domain dependencies.
We will then present methods that exploit the same type of assumptions, and how we take a different interpretation of these assumptions.

\subsection{Manifold Assumption}
According to the manifold assumption, high-dimensional data can be embedded in a lower-dimensional space \cite{SSL_book}, a latent space, with a lower degree of freedom considered. For the algorithm, working with this latent code representation of the data is easier than working with the original high-dimensional data, and learning this latent space helps to capture the underlying data structure - the intra-domain dependencies.
Several architectures such as the auto-encoder and variational auto-encoder have been used to learn this latent space \cite{ae,ae1,ae2,vae}.
While learning intra-domain dependencies has shown effectiveness, allowing to use of additional data without associated views in the other domains \cite{IODA}, this approach alone does not leverage inter-domain dependencies when there is no supervision. The next assumption help to alleviate this issue.

\subsection{Shared Latent Space Assumption}
Deriving from the Manifold Assumption and mainly used in DA, the Shared Latent Space Assumption state that associated views of different domains can be mapped to a common latent code. If a view $d_i\sim D_i$ is associated with view $d_j \sim D_j$, then $\tilde d_i = \tilde d_j$.
Learning to align each manifold in the same shared structure allows learning the inter-domain dependency between views of different domains.
But this every-to-one mapping can be seen as an over-simplification, as multiple modalities can have very dissimilar features (e.g. consider an image and its pixel-wise semantic segmentation mask,  the mask does not contain in itself information about the texture), this inter-domain discrepancy coupled with the every-to-one mapping could hinder the model when learning the translation.

\begin{figure}%
    \centering
    \subfloat[\centering A single shared latent space]{{\begin{tikzpicture}
  
  \node[ellipse,minimum width=15mm,minimum height=10mm, draw=cyan,fill=cyan!50] (a) at (0,0) {$A$};
  \node[ellipse,minimum width=10mm,draw=black,fill=black!10,right=5mm of a] (z) {$Z$};
  \node[ellipse,minimum width=15mm,minimum height=10mm,draw=yellow!90!black,fill=yellow!30,right=5mm of z] (b) {$B$};
  
  \draw[->] (a)  to[bend left=45] (z);
  \draw[->] (z) to[bend left=45] (b);

  \draw[->] (b)  to[bend left=45] (z);
  \draw[->] (z) to[bend left=45] (a);

\end{tikzpicture}}}%
    \qquad
    \subfloat[\centering Multiple latent spaces space]{{\begin{tikzpicture}
  
  \node[ellipse,minimum width=15mm,minimum height=10mm, draw=cyan,fill=cyan!50] (a) at (0,0) {$A$};
  \node[ellipse,minimum width=10mm,draw=black,fill=black!10,right=5mm of a] (za) {$Z_A$};
  \node[ellipse,minimum width=10mm,draw=black,fill=black!10,right=5mm of za] (zb) {$Z_B$};
  \node[ellipse,minimum width=15mm,minimum height=10mm,draw=yellow!90!black,fill=yellow!30,right=5mm of zb] (b) {$B$};
  
  \draw[->] (a)  to[bend left=45] (za);
  \draw[->] (za) to[bend left=45] (zb);
  \draw[->] (zb) to[bend left=45] (b);

  \draw[->] (b)  to[bend left=45] (zb);
  \draw[->] (zb) to[bend left=45] (za);
  \draw[->] (za) to[bend left=45] (a);

\end{tikzpicture}}}%
    \caption{Two types of latent space are commonly used in domain translation
    (a) a single shared latent space,
    (b) or using multiple latent spaces}%
    \label{fig:related_work:latent_space}%
\end{figure}
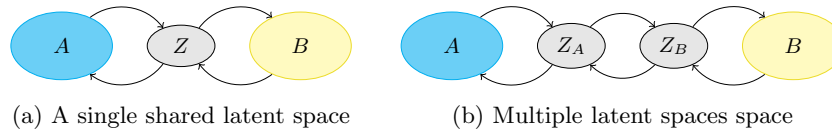

\subsection{Domain Translation}
\label{related_works:domain_translation}
We will give an overview of common domain translation methods while classifying them according to the previously defined assumptions in order to better understand how we take different choices from them.
These models can be divided into roughly two types illustrated in Figure~\ref{fig:related_work:latent_space}.

Methods using a \textbf{single shared latent} space can use adversarial learning on the feature space \cite{GRL,shared1corrnet,shared2alignflow,shared3,shared4}, and possibly also on the original high-dimensional space \cite{unit,pix2pix,stargan,unpairedGAN,shared2alignflow,dcan,hoffman2017cycada,Li_2019_CVPR}, using generative adversarial networks (GAN \cite{GAN}) to generate a view from a latent code. GAN has been notoriously hard to train due to the min-max optimization that can lead to vanishing gradient \cite{wasserstein} and is subject to mode collapse~\cite{collapse} (a mode collapse occurs when the generation is almost deterministic and the generator always creates the same output).
Building on top of these existing works, models with \textbf{multiple latent spaces} often use disentanglement learning, dividing the latent code into two-part: a domain-agnostic content code, and a domain-specific style code. This more sophisticated setting allows learning a domain translation function that would combine a content code from a source domain and the style code of the targeted domain \cite{drit,munit,unpairedGAN,sun2019dualglow,liu2020gmmunit,vu2018domain}.
Learning a disentangled representation is not straightforward as the definition in itself of content and style is ambiguous. These approaches bring great results at the cost of additional complexity and inductive bias (e.g. using adversarial learning between the content code and the style code).

%%%%%%%%%%%%%%%%%%%%%%%%%%%%%%%%
\subsection{Latent Space Mapping Assumption}
We will now present how \model uses the previously defined concept and how it differs from existing works by taking a different interpretation of the presented assumptions.
\model use the manifold assumption by, for each domain $D_i$, learning a latent space where a view $d_i$ is embedded in its latent code $\tilde d_i$. This allows the approach to modelize intra-domain dependencies.
Additionally, we learn the inter-domain dependencies using the shared latent space assumption, but here we decide to take a different approach from the classical full-shared latent space while keeping its simplicity.
Instead of using a single shared latent space, we relax the assumption by supposing the existence of a function $L_{i\rightarrow j}: \tilde D_i \rightarrow \tilde D_j$ allowing to translate a latent code from one domain to the latent code of another domain.
The shared assumption is then applied between translated codes $L_{i\rightarrow j}(\tilde d_i)$ and codes $\tilde d_j$ from the original domain.
Similar ideas has been approached with success in existing works: in IODA~\cite{IODA} by the mean of pre-training an input encoder using input data, an output decoder using output data to initialize a network, and learning the final task in a supervised fashion.
And in SOP~\cite{SOP} where there is no pre-training but the learning of this latent space is done in an unsupervised fashion as a regularization task. But SOP only uses the manifold assumption and does not take into account the discrepancy between a translated latent code and a latent code from the original domain. To alleviate this issue, we will use a framework similar to that of the SOP. We further regularize each latent space by making the assumption that $L_{i\rightarrow j}(\tilde d_i)$ should be close to $\tilde d_j$.
\section{\model Framework}
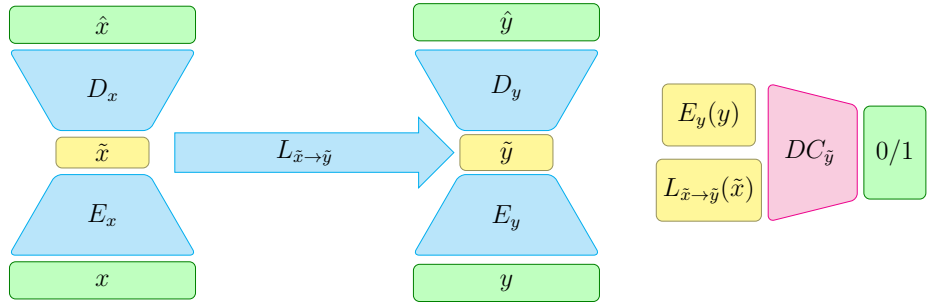
\begin{figure}
\centering
\begin{tikzpicture}
\tikzstyle{every node}=[font=\large]
  
\node[rectangle,rounded corners=1mm, minimum width=15mm, draw=yellow!50!black, fill=yellow!50!white] (xt) at (0,0) {$\tilde{x}$};
\node[rectangle,rounded corners=1mm,minimum width=30mm, minimum height= 6mm, draw=green!50!black, fill=green!25!white, below= 15mm of xt] (x) {$x$};
\node[rectangle,rounded corners=1mm,minimum width=30mm, minimum height= 6mm, draw=green!50!black, fill=green!25!white, above= 15mm of xt] (xe) {$\hat{x}$};

\node[rectangle,rounded corners=1mm, minimum width=15mm, draw=yellow!50!black, fill=yellow!50!white, right = 50mm of xt] (yt) {$\tilde{y}$};
\node[rectangle,rounded corners=1mm,minimum width=30mm, minimum height= 6mm, draw=green!50!black, fill=green!25!white, below= 15mm of yt] (y) {$y$};
\node[rectangle,rounded corners=1mm,minimum width=30mm, minimum height= 6mm, draw=green!50!black, fill=green!25!white, above= 15mm of yt] (ye) {$\hat{y}$};

\path (xt.east)--(yt.west) node[pos=0.5,single arrow, draw=cyan, fill=cyan!25!white, minimum height=45mm] (lxy) {$L_{\tilde{x}\rightarrow\tilde{y}}$};

\coordinate[above=1mm of x.north] (xn);
\coordinate[below=1mm of xt.south] (xts);

\draw[draw=cyan, fill=cyan!25!white, rounded corners=1mm] (x.west|-xn)--(x.east|-xn)--(xt.east|-xts)--(xt.west|-xts)--cycle;
\path (x) -- (xt) node[pos=0.5] (ex) {$E_x$};

\draw[draw=cyan, fill=cyan!25!white, rounded corners=1mm] (y.west|-xn)--(y.east|-xn)--(yt.east|-xts)--(yt.west|-xts)--cycle;
\path (y) -- (yt) node[pos=0.5] (ey) {$E_y$};

\coordinate[above=1mm of xt.north] (xtn);
\coordinate[below=1mm of xe.south] (xes);

\draw[draw=cyan, fill=cyan!25!white, rounded corners=1mm] (xe.west|-xes)--(xe.east|-xes)--(xt.east|-xtn)--(xt.west|-xtn)--cycle;
\path (xt) -- (xe) node[pos=0.5] (ex) {$D_x$};

\draw[draw=cyan, fill=cyan!25!white, rounded corners=1mm] (ye.west|-xes)--(ye.east|-xes)--(yt.east|-xtn)--(yt.west|-xtn)--cycle;
\path (yt) -- (ye) node[pos=0.5] (ey) {$D_y$};

%%%-----

\node[rectangle,rounded corners=1mm,minimum width=10mm, minimum height=15mm, draw=green!50!black, fill=green!25!white, right=50mm of yt] (disout) {$0/1$};

\path (yt)--(disout)--+(-30mm,00)coordinate(disin) {};

\node[rectangle,rounded corners=1mm,minimum width=15mm, minimum height=10mm, draw=yellow!50!black, fill=yellow!50!white, below=1mm of disin] (disinlxy) {$L_{\tilde{x}\rightarrow\tilde{y}}(\tilde{x})$};
\node[rectangle,rounded corners=1mm,minimum width=15mm, minimum height=10mm, draw=yellow!50!black, fill=yellow!50!white, above=1mm of disin] (disiney) {$E_y(y)$};

\coordinate[right=2mm of disiney.east] (disine);
\coordinate[left=1mm of disout.west] (disoutw);

\draw[draw=magenta, fill=magenta!25!white, rounded corners=1mm] (disine|-disinlxy.south)--(disine|-disiney.north)--(disoutw|-disout.north)--(disoutw|-disout.south)--cycle;
\path (disin-|disiney.east) -- (disout.west) node[pos=0.5] (ey) {$DC_{\tilde{y}}$};

\end{tikzpicture}
\caption{Illustration of \model framework. The translation is presented from the point of view of $X\rightarrow Y$}
\label{fig:Model_illustration}
\end{figure}

We now present \model framework, illustrated in Figure~\ref{fig:Model_illustration}, for domain translation between multiple modalities and multiple domains, by learning functions $f_{D_i \rightarrow D_j}: D_i\rightarrow D_j$.
The framework learns intra-domain dependencies, for each view $d_i \sim D_i$, the model produces its embedding $\tilde d_i$, this can be done in an unsupervised way by the mean of reconstruction task.
It also relaxes the shared latent space assumption by supposing the existence of a function $L_{i \rightarrow j}$ allowing to translate of a latent code $\tilde d_i$  to another domain latent space $\tilde d_j$, the shared assumption is then applied after the translation by the mean of a distance loss when paired information is available or by means of adversarial in an unpaired setting.

For clarity of concern and without loss of generality, we will now consider only two domains $X$ and $Y$, and we will present the translation from domain $X$ to domain $Y$, the same losses are applied when translating from $Y$ to $X$. The under-script $x\rightarrow y$ will be used on functions that are specific to the translation of $X$ to $Y$ and has to be duplicated in the case of $Y$ to $X$. Other functions are only applied once.
This approach leverage three types of data configuration, either learning from view $x \sim \px$ when $y$ is not available, from $y \sim \py$ when it is provided without corresponding $x$, or from paired views $(x,y) \sim \pxy$.
Our goal is to learn a translation function $f_{x \rightarrow y}:X \rightarrow Y$ from domain $X$ to domain $Y$ as $\hat y=f_{x \rightarrow y}(x)$.
Using the manifold assumption we introduce two latent spaces $\tilde x$ (resp. $\tilde y$) where the view $x$ (resp. $y$) could be encoded and where the code would retain a summarized version of the original data.
To learn this latent space, two auto-encoders are used: $\tilde x = E_x(x)$ and $\tilde y = E_y(y)$ mapping a view to its latent code, and $\hat x=D_x(\tilde x)$ and $\hat y=D_y(\tilde y)$ mapping the latent code to its original view. An additional function performs the mapping between latent space $X$ to $Y$ as $\tilde y = \Lxy(\tilde x)$.
Given these latent spaces and functions, $f_{x \rightarrow y}$ can be reformulated as: $\hat y=f_{x \rightarrow y}(x)=D_y(\tilde y)=D_y \circ \Lxy(\tilde x)=D_y \circ\Lxy\circ E_x(x)$.

\subsection{Learning $X$ to $Y$ Domain Translation}
\label{part:translation_xy}

When a pair is available as $(x,y)\sim \pxy$, the model performs as a fully supervised method, predicting $y$ from $x$.
\begin{align}
    \hat y = D_y \circ \Lxy  \circ E_x(x)
\end{align}
This allows learning the translation network by optimizing the supervised loss
\begin{align}
\label{loss:xy}
    \min_{\theta_{E_x},\theta_{\Lxy},\theta_{D_y}}
    \mathbb E_{\pxy} \mathcal L_{x \rightarrow y}
    (D_y \circ \Lxy \circ E_x(x), y) = \mathcal L_{x \rightarrow y}(\hat y, y)
\end{align}
Where $\mathcal L_{x \rightarrow y}$ can be a cross-entropy loss for a segmentation problem, or a MSE loss for a regression problem for example.

\subsection{Intra-Domain Dependencies Regularization}
\label{model:intra_domain_dependencies_regularization}
When a pair of views is available as $(x,y)\sim \pxy$, or when only $x \sim \px$, or only $y \sim \py$ are available, it is possible to regularize the training of the model by adding two additional reconstruction losses. 
These losses based on the manifold assumption allow to learn intra-domain dependencies via the mean of reconstruction, with auto-encoder for example \cite{ae,ae1,ae2}.
The model optimizes the reconstruction of the views from their latent code as:
\begin{align}
    \hat x = D_x(\tilde x) = D_x \circ E_x (x)
    \\
    \hat y = D_y(\tilde y) = D_y \circ E_y (y) 
\end{align}
The two associated reconstruction losses
\begin{align}
    \min_{\theta_{E_x},\theta_{D_x}}
    \mathbb E_{\px} \mathcal L_{x \rightarrow x}(D_x \circ E_x(x), x) 
    = \mathcal L_{x \rightarrow x}(\hat x, x)
    \\
    \min_{\theta_{E_y},\theta_{D_y}}\mathbb E_{\py} \mathcal L_{y \rightarrow y}(D_y \circ E_y(y), y) =\mathcal L_{y \rightarrow y}(\hat y, y)
\end{align}
can be used also when not all the data are available. When only $x$ (resp. $y$) is available, only $\mathcal L_{x \rightarrow x}$ (resp. $\mathcal L_{y \rightarrow y}$) is optimized.

In machine learning, it is a common case to have plenty of unlabeled $x$ only views, but to have only labels without the $x$ available is more rare.
For example, in a domain translation as style transfer, having $y$ only sample is often easy to obtain as it needs collecting images of the desired style.
For some other tasks like semantic segmentation, it is possible to obtain $y$ only sample in road segmentation through the usage of simulator \cite{GTA}, in medical images segmentation the use of phantom able the generation of $y$ only sample \cite{XCAT}.

So far, by only taking the translation from $X$ to $Y$, this framework corresponds to the SOP architecture~\cite{SOP}. \model extends SOP in order to allow multi-domain translation by taking into account the transition from $Y$ to $X$ and further leveraging inter-domain dependencies.

\subsection{Inter-Domain Dependencies Regularization}
Unlike previous approaches using the shared latent space assumption, we consider that this assumption should be relaxed and that $\Lxy(\tilde x)$ should be close to $\tilde y$.
Applying inter-domain regularization on a fully shared latent space between each latent code from different views without domain-specific information might hinder the learning process by removing uncorrelated, domain-specific features which might be helpful for the domain translation task.
\model prevents this issue by creating a latent space for each domain, and regularization is done between a latent code 
 $\tilde y$ originating from a domain and a translated latent code into this domain $\Lxy(\tilde x)$. This does not penalize keeping domain-specific information in each latent code.

To enforce latent code $\Lxy(\tilde x)$ to be close to $\tilde y$, two additional losses are added, depending on the available supervision.

\paragraph{Supervised case} In the case where the pair of view $(x,y)$ is available, \model minimizes
\begin{align}
\label{loss:distance}
    \min_{\theta_{E_x},\theta_{\Lxy},\theta_{E_y}}
    \mathbb E_{\ptxy} 
    \lossdist(\tilde x,\tilde y)=d(\Lxy(\tilde x), \tilde y)
\end{align}
where $d$ is a distance function, as for example $L_1$ distance.
Minimizing $\lossdist$ allows bringing closer the translated latent code $\Lxy(\tilde x)$ and the latent code $\tilde y$.
In the supervised case, where all views are available, for example in a road semantic segmentation context considering the RGB image and its segmentation map $(x, y)$, and their latent representation $(\tilde x, \tilde y)$.
Supposing that $\tilde y$ allows reconstructing $y$ through the decoder $D_y(\tilde y)=y$, then if $\Lxy(\tilde x) = \tilde y$, obtaining $\tilde y$ using $x$ would allow to translate $x$ into $y$ through the latent space mapping.
In order for the translation to be accurate, one would expect $\Lxy(\tilde x)$ to be close to $E_y(y) = \tilde y$.
Equation \ref{loss:distance} enforce this constraint by minimizing the distance between latent codes.

\paragraph{Unsupervised case}  When the supervision is incomplete (either only $x$ or only $y$ is available), we substitute the distance loss $\lossdist$ with a feature adversarial loss inspired by existing domain adaptation approaches as done in \cite{cite_confusion_loss}.

We would like the translated latent code $\Lxy(\tilde x)$ to be invariant from the latent code $\tilde y$. Classical machine learning models assume that all the input came from the same distribution, in this case, the model is the decoder, and the inputs are latent codes.
We consider latent codes to be invariant if a domain classifier is unable to distinguish its source.
For that purpose, a domain classifier $\dcy$ is added on top of $Y$ latent space in order to draw the two distributions closer. For the domain classifier, the loss is a binary cross-entropy (BCE) where the discriminator has to correctly classify the presented latent code in the right domain.
\begin{align}
\label{loss:critic}
    \min_{\theta_{\dcy}} \mathbb{E}_{\ptx, \pty}
    \mathcal L_{\dcy}(\tilde x,\tilde y)=
    \text{BCE}(\dcy\circ \Lxy(\tilde x),0) + \text{BCE}(\dcy(\tilde y), 1) 
\end{align}
The domain classifier attempt to discern from which domain the presented latent code is sampled.

Classical adversarial approaches make the generating part maximize the error of $\dcy$, however, as stated in \cite{cite_confusion_loss2}, the two distributions $\Lxy(\tilde x)$ and $\tilde y$ are changing, this could lead to an optimization problem. When the generating part is optimal, the domain classifier would only have to flip its sign in order to produce the correct prediction. This would result in oscillations in the optimization.
Therefore, we encourage domains to produce features $\Lxy(\tilde x)$ and $\tilde y$ to be indistinguishable, using a confusion loss \cite{cite_confusion_loss}, where we compare the domain classifier decision against a uniform distribution:
\begin{align}
    \min_{\theta_{E_x},\theta_{E_y},\theta_{\Lxy}}
    \mathbb{E}_{\ptx, \pty}
    \lossconf(\tilde x, \tilde y)
    =&
    \text{BCE}\left(\dcy\circ \Lxy(\tilde x),\frac12\right)\\
    &+
    \text{BCE}\left(\dcy(\tilde y), \frac12\right)
\end{align}

\subsection{Final Loss}
Finally, the model is trained to optimize the weighted sum of all the previous losses. For a two-way $X$/$Y$ domain translation ($x\rightarrow y$ and $y \rightarrow x$), the global loss is
\begin{align}
    \min_{\substack{\theta_{E_x},\theta_{E_y},\theta_{\Lxy},\\ \theta_{D_x},\theta_{D_y},\theta_{\Lxyx}}}&
    \mathcal{L}_{Final}=
    \lambda_1 \mathcal L_{x \rightarrow x}
    +\lambda_2 \mathcal L_{y \rightarrow y}
    +\lambda_{3}^1 \mathcal L_{x \rightarrow y}
    +\lambda_{3}^2 \mathcal L_{y \rightarrow x} \nonumber \\
    &+\lambda_{4} (\lossdistx + \lossdist)
    +\lambda_{5} (\lossconfx + \lossconf) \\
    \min_{\theta_{\dcx}, \theta_{\dcy}} &
    \mathcal{L}_{Adversarial}=
    \mathcal L_{\dcx}+\mathcal L_{\dcy}
\end{align}
It allows the model to receive supervision regardless of whether the samples are drawn from a marginal distribution $x\sim \px$ and $y \sim \py$, or from their joint distribution $x,y \sim \pxy$. $\lossdist$ and $\lossconf$ applied on a latent space on domain $Y$ further push the model to create an embedding where the translated latent code $\Lxy(\tilde x)$ and the original latent code $\tilde y$ are more similar.
In the general cases where all domains are similar (e.g. both domains are real images), $\lambda_{3}^1=\lambda_{3}^2$.
\section{Experiments}
\label{partie:Experiments}

In this section, we provide an evaluation of \model along with three baselines on three datasets.
To evaluate the benefits induced by \model, we evaluate every approach using the same architecture that we made as simple as possible.
We empirically show that the use of \model is not detrimental when a lot of supervision is available. Furthermore, \model allows for improving the training when additional unsupervised information is available and improves the training while reducing the variance of the models.
Qualitative results are available in Appendix~\ref{app:qualitative}.

\subsection{Datasets}
\label{exp:datasets}
\begin{figure}
\centering
\includegraphics[height=4cm]{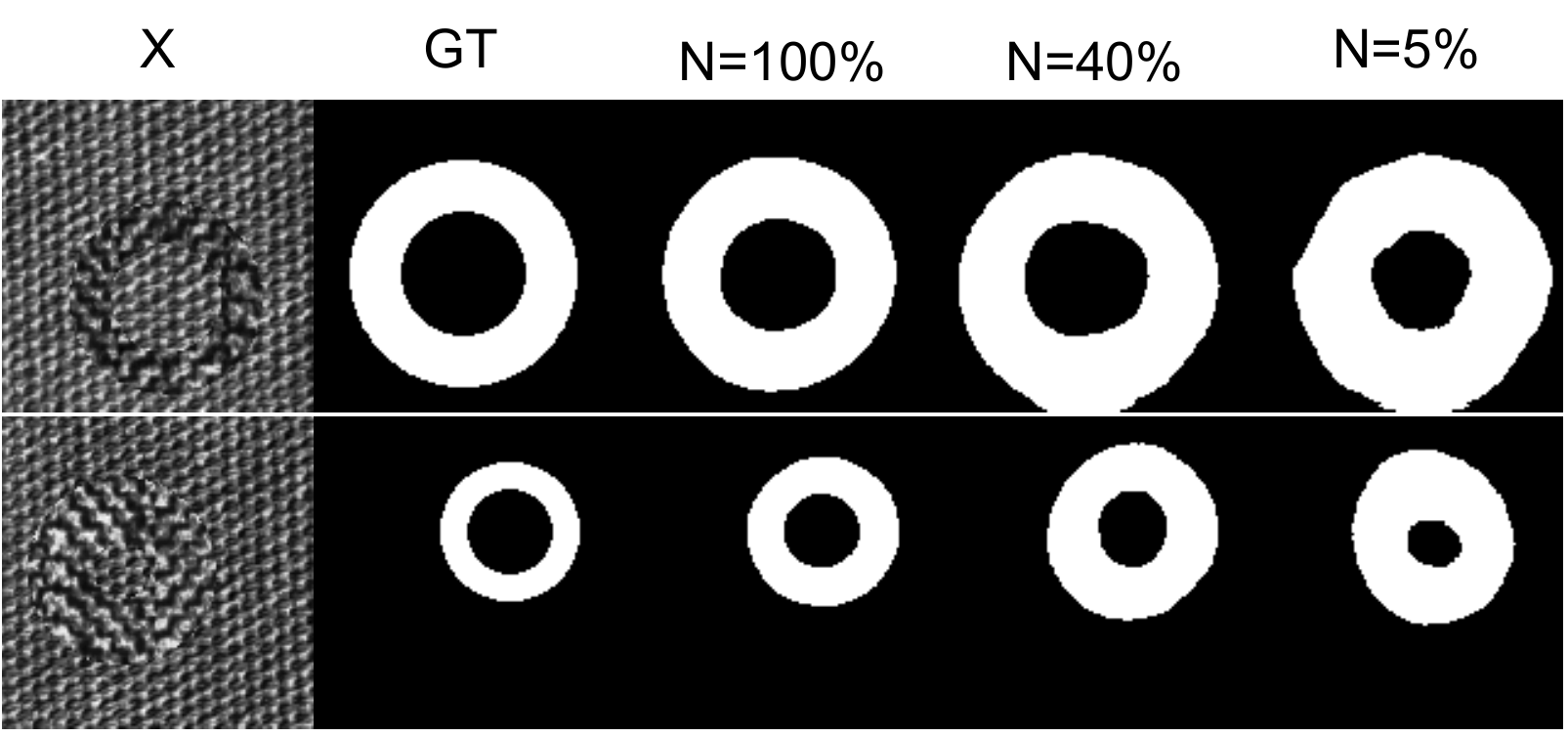}
\caption{\toyS samples with their associated prediction from \model for different amounts of available $(x,y)$ views. More predictions in Appendix~\ref{app:qualitative}}
\label{fig:toyset_dataset}
\end{figure}

\subsubsection{\TOY (\toyS) Dataset}\label{exp:datasets:toy}
The \toy dataset consists of image pairs made by superposing a background texture and a ring from another texture \footnote{https://www.ux.uis.no/$\sim$tranden/brodatz.html} for the input domain and the associated circle segmentation for the output domain.
Four parameters control the texture generation: circle x position, circle y position, circle radii, and circle thickness. In order to complexify the task, for the segmentation mask, we swapped the x position with the circle thickness and the y position with the circle radii. This changes the task from a segmentation task to a more challenging image-to-image translation task.
We use the standard mean Intersection over Union (mIoU) as evaluation metric for the output prediction.
Figure~\ref{fig:toyset_dataset} displays two pairs of an input image and ground truth. We refer to Appendix~\ref{app:dataset_details:toyset} for dataset description details.

\subsubsection{Facial Landmark Detection (\faceS) Dataset}
\begin{figure}
\centering
\includegraphics[height=6cm]{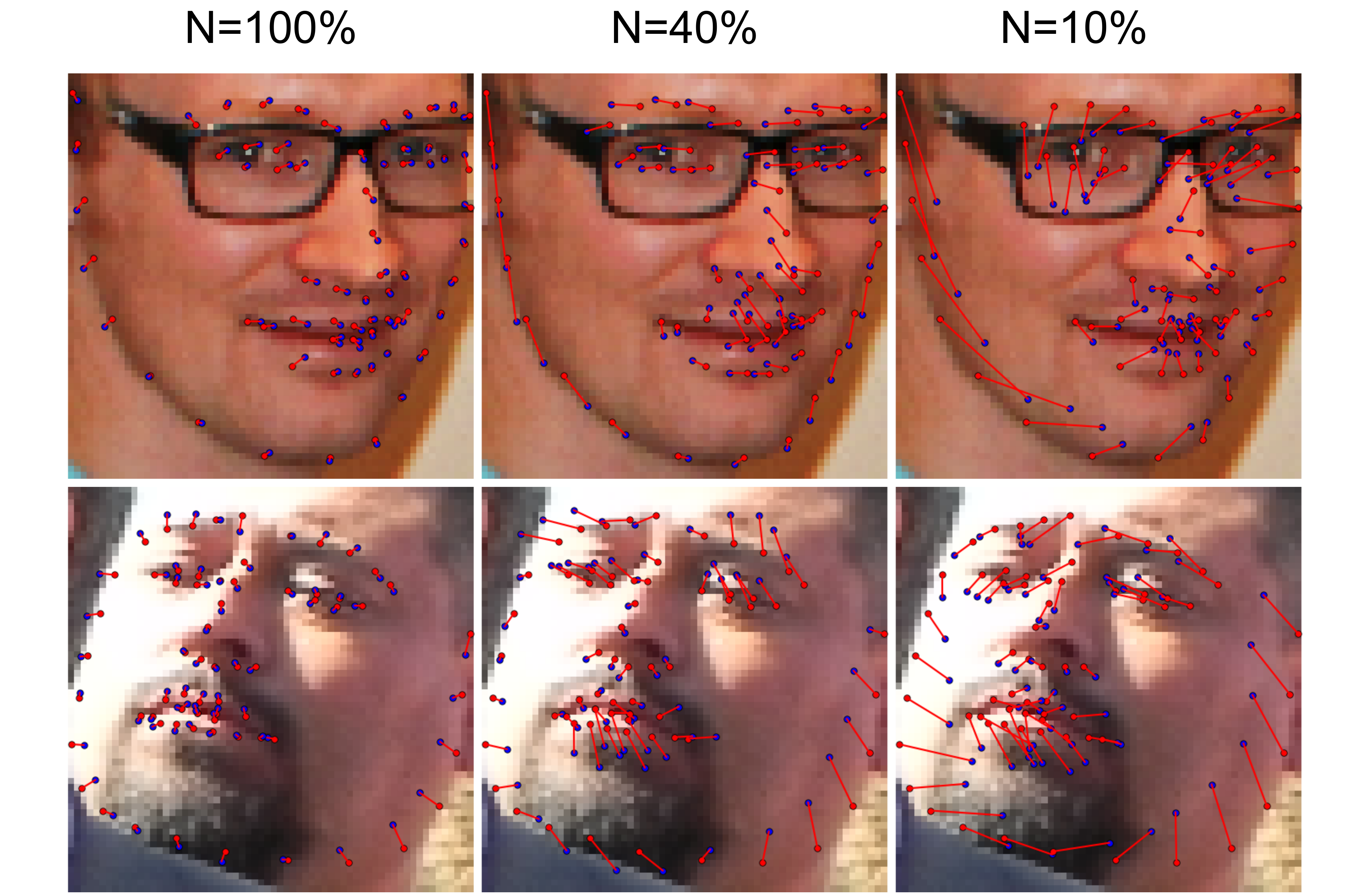}
\caption{\faceS with their associated facial landmarks (red dot) and predicted landmark from \model (blue dot) for different amounts of available $(x,y)$ views. More predictions in Appendix~\ref{app:qualitative}}
\label{fig:face_dataset}
\end{figure}

We study the real case of facial landmark detection, on the 300 Faces In-the-Wild Challenge (\faceS) \cite{Face_dataset_1,Face_dataset_2}. This 300 Faces In-the-Wild Challenge dataset is constituted of 300 indoor images and 300 outdoor images with associated 68 facial landmarks for each face. It provides a context with multiple modalities (image faces, and vector of positions) and with a highly structured output domain that presents strong intra-dependencies.
We use the standard Normalized Root Mean Square Error (NRMSE) as metric for facial landmark detection.
We refer to Appendix~\ref{app:dataset_details:face} and \cite{Face_dataset_1,Face_dataset_2} for dataset description details, examples of data are given in Figure~\ref{fig:face_dataset}.

\subsubsection{Sarcopenia Semantic Segmentation Dataset  (\sarcoS)}
\begin{figure}
\centering
\includegraphics[width=0.9\textwidth]
{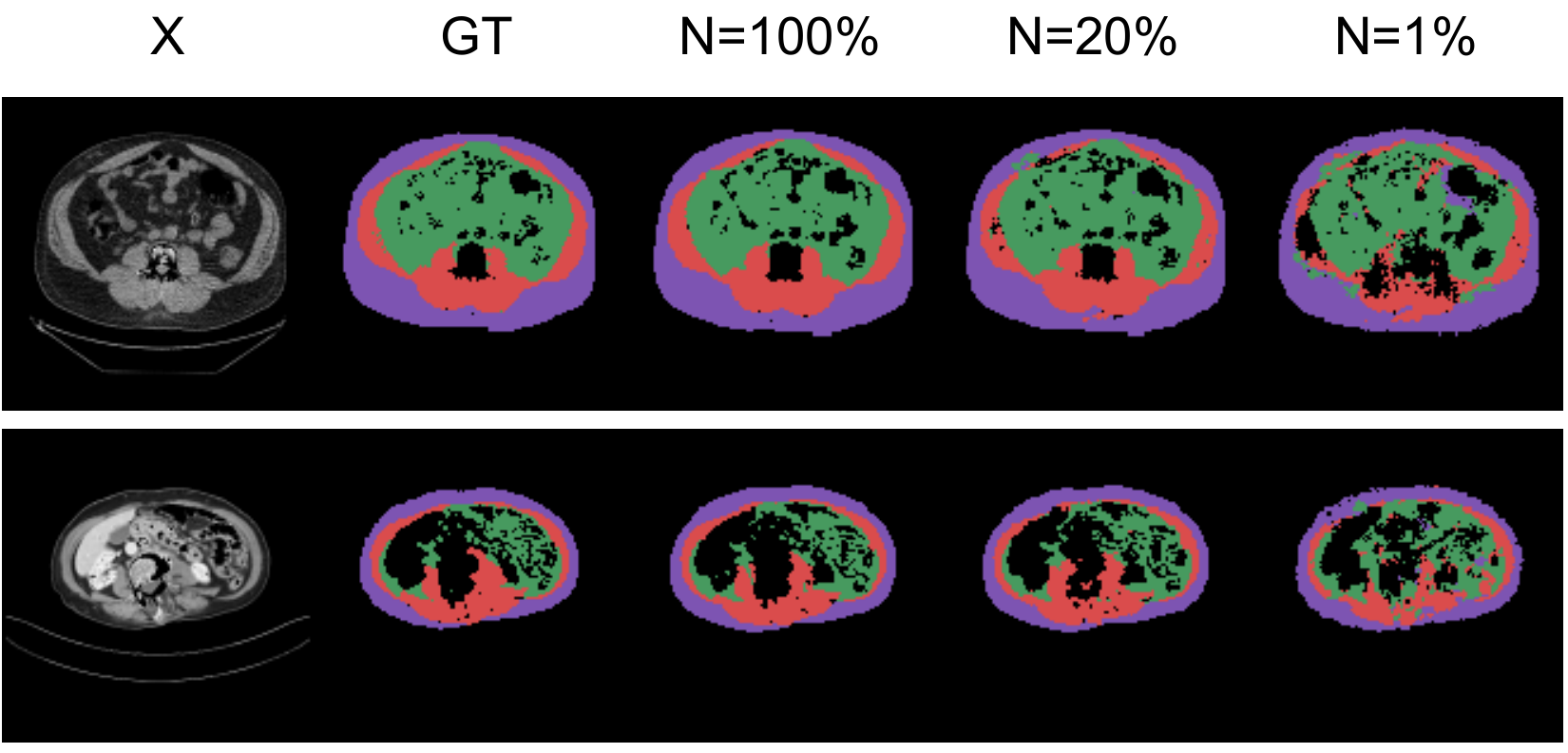}
\caption{\sarcoS samples with their associated prediction from \model for different amounts of available $(x,y)$ views. More predictions in Appendix~\ref{app:qualitative}}
\label{fig:sarcopenia_dataset}
\end{figure}

We study segmentation of the third lumbar vertebra (L3) level from Computed Tomography (CT) scans. This task is needed for predicting the sarcopenia level, which is a measure of muscle atrophy, later used in cancer diagnostic \cite{sarco_cancer_1,sarco_cancer_2}.
We use the mean Intersection over Union (mIoU) as metric for the segmentation.
We refer to Appendix~\ref{app:dataset_details:sarcopenia} for dataset description details, examples of data are given in Figure~\ref{fig:sarcopenia_dataset}.

\subsection{Baselines}
We compare our model with 5 baselines: a basic baseline, IODA~\cite{IODA}, SOP~\cite{SOP}, CycleGAN~\cite{cyclegan} an unsupervised domain translation approach, and Pix2Pix~\cite{pix2pix} a supervised domain translation approach. Every baseline used has the same model architecture.
More details on the implementation can be found in Appendix~\ref{app:exp_det}.

The \textbf{basic baseline}, is restricted to $D_y \circ \Lxy \circ E_x$ as in Section~\ref{part:translation_xy}, this corresponds to using only the input to output translation with the loss described in equation~\ref{loss:xy}.
%\subsubsection{IODA} \cite{IODA}
\textbf{IODA}~\cite{IODA} model has a similar structure as our model and also exploits information from input data and output data, but the learning of additional information is performed offline during a pre-training phase. Namely IODA first pre-trains $D_x \circ E_x$ and $D_y \circ E_y$. Then in a final phase, the model $D_y \circ \Lxy \circ E_x$ is initialized using the pre-trained weights and training using equation~\ref{loss:xy}.
\textbf{SOP}~\cite{SOP} has a similar architecture, and uses information from input and output as a mean to regularize the training of equation~\ref{loss:xy}. The difference between the SOP model and our model lies in the equation~\ref{loss:distance} and equation~\ref{loss:critic} that are not present in SOP.
Namely, a distance cost brings closer representations from different origins, and an adversarial cost pushes closer different embedding to follow the same distribution.
\textbf{CycleGAN}~\cite{cyclegan} learns domain translation in an unsupervised way. Unlike IODA, SOP, and \model, it does not explicitly modelizes a latent space. The translation $x \rightarrow y$ is learned by a generator $D_y \circ \Lxy \circ E_x$ and is enforced realistic using a domain discriminator, a cycle consistency further regularizes the translation.
\textbf{Pix2Pix}~\cite{pix2pix} learns domain translation in a supervised way. Not explicitly modelizing a latent space, Pix2Pix learns a generator $D_y \circ \Lxy \circ E_x$. The supervision came from a discriminator that takes as input either $(x,y)$ as the real examples or $(x, D_y \circ \Lxy \circ E_x (y))$ as the fakes examples, with an additional supervised loss between $y$ and $D_y \circ \Lxy \circ E_x (y)$.

\subsection{Adapting to Lower Supervision}
\label{partie:lower_supervision}

% double figure
\begin{figure}%
    \centering
    \subfloat[\centering]
    {\includegraphics[width=10cm]
    {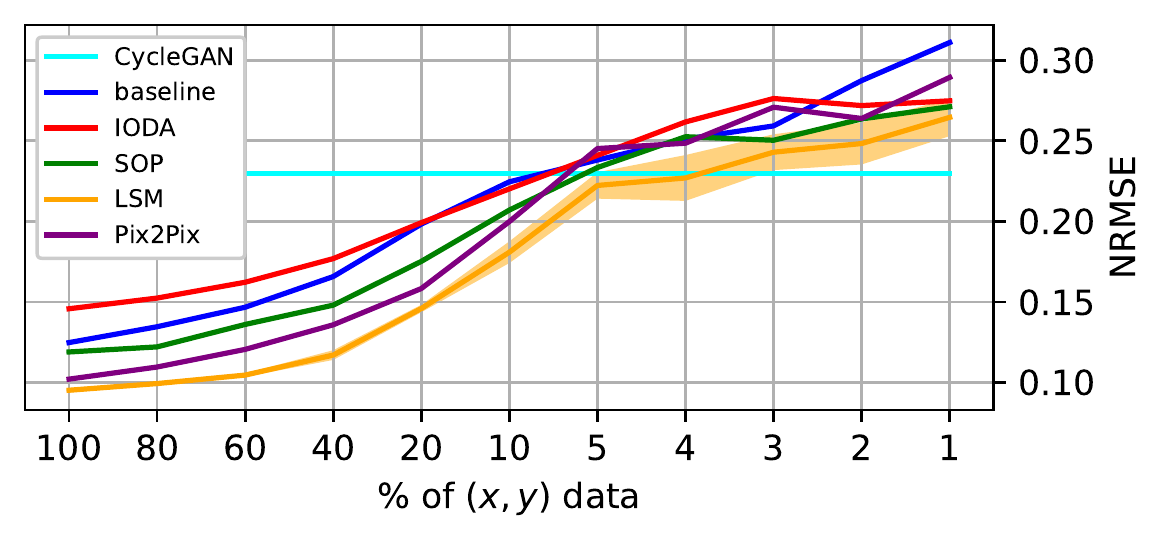}\label{fig:face:lower_supervision}}%
    \qquad
    \subfloat[\centering]
    {\includegraphics[width=10cm]
    {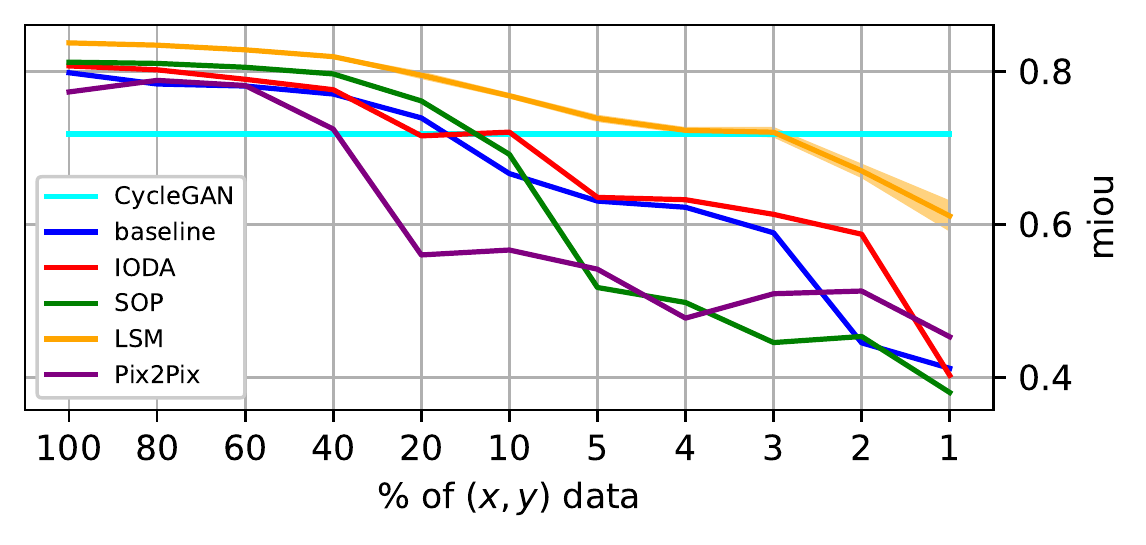}\label{fig:sarcopenia:lower_supervision}}%
    \caption{Impact of the number of available pairs from (a) the \faceS dataset on the NRMSE metric (lower is better) and (b) the \sarcoS dataset on the mIoU metric (higher is better). Areas around \model line represent the standard deviation (5 trained models for each baseline)}%
\end{figure}

\begin{figure}
\centering
\includegraphics[width=10cm]
{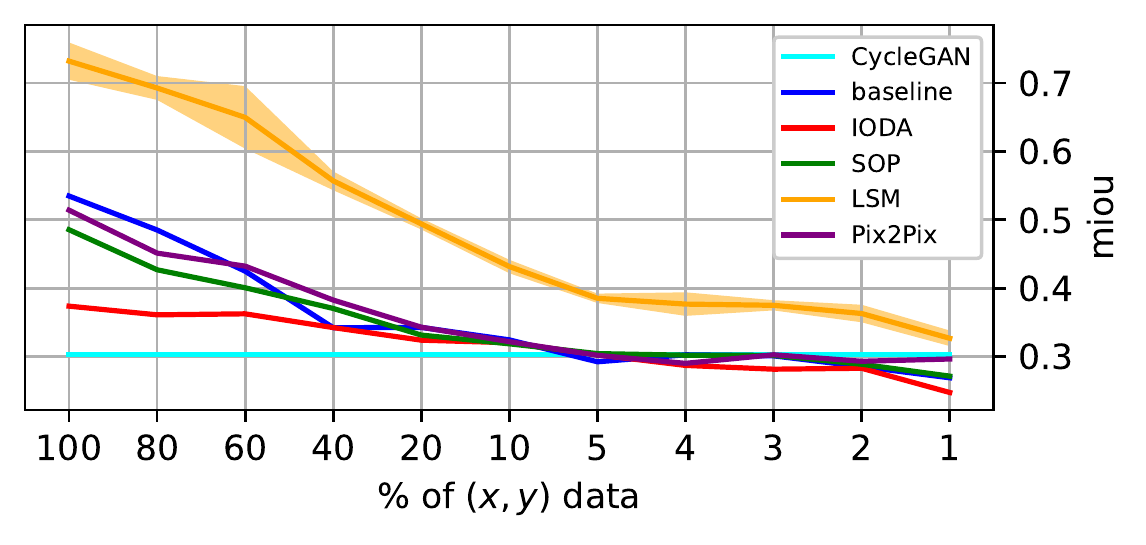}
\caption{Impact of the number of available pairs from the \toy dataset on the mIoU metric (higher is better). Areas around \model line represent the standard deviation (5 trained models for each baseline)}
\label{fig:toy:lower_supervision}
\end{figure}

% setup explanation
We first compare each model while adapting the amount of available supervision, controlled by taking $N$\% of the dataset as supervised pair $(x,y)$, and splitting evenly the remaining $(100-N)$\% as only $x$ samples and only $y$ samples (unsupervised pairs). We lower $N$ in order to see an evolution in the metrics.
To ensure that no $x$ and $y$ in the unsupervised set could come from the same pair, we divide the dataset in two before the training, keeping one part as a bank of unsupervised pairs. This has for effect to divide the effective dataset length by two.
As CycleGAN operates in an unsupervised setting, its performance remains the same for each $N$ value.
\model is compared for different values of $N$ for the facial landmark in Figure~\ref{fig:face:lower_supervision}, the sarcopenia dataset in Figure~\ref{fig:sarcopenia:lower_supervision} and the synthetic image translation in Figure~\ref{fig:toy:lower_supervision}.

\paragraph{High Amount of Supervision}
For the highest $N$ values ($N=100$) and the \faceS and \sarcoS dataset, there are no significant differences between models, as there is plenty of $(x,y)$ views, the different baselines are able to learn the tasks. For \toyS we found that \model greatly improves on the metric compared to the baselines, this is explained by the fact that, unlike in the two previous tasks, there is no direct pixel-to-pixel (or spatial) correspondence between the two domains (due to the swapping of variables).
We found that CycleGAN is able to perform on the \faceS and \sarcoS tasks, where there is a pixel-to-pixel correspondence, but fails the \toyS image translation, as the correlation between domains is not pixel-to-pixel, this unsupervised approach is not able to understand the swapping of variables and not suited for this task.

Medium values of $N$($N\in \{80,60,40\}$ for facial landmark, $N\in \{80, 60, 40\}$ for sarcopenia datasets, $N\in \{80\}$ for synthetic dataset) only slightly affect the training. For the two first, the metric only starts to really decrease for \model and the baselines when $N<40$, this is explained by the fact that the models does not need every sample to learn the task. Those values of $N$ are Pareto optimal for this combination of dataset and model leading to either no or a slight decrease in the metrics. For the synthetic dataset \toyS, the metric start to decrease earlier as the task might be harder for the models and the amount of data is lower.

\paragraph{Low Amount of Supervision}
When $N$ became low ($N<60$ for every dataset) and the amount of supervision became limited, every model's performance start to degrade as the amount of supervision is no longer sufficient to completely learn the task and generalize over the test set.

\begin{table}
\centering
\caption{NRMSE evaluation on the Facial Landmark Dataset according to the portion of available pairs $(x,y)$ from the dataset}
\label{table:exp:lower_supervision:face}
\includegraphics[width=\textwidth]{table_data/face.tex}
\end{table}

For the \faceS and \toyS datasets (Figure~\ref{fig:face:lower_supervision})  and the lowest amount of supervision ($N=1$), \model performances  decline and reach the same NRMSE as the supervised and unsupervised baselines. While for the \sarcoS dataset, \model is able to better adapt to a low amount of supervision than the other supervised baselines.
In the low-supervision setting, unsupervised models such as CycleGAN are able to perform well, for $N\leq4$ for \faceS, and $N\leq2$ for \sarcoS. For the synthetic dataset, the task is impossible to solve without supervision due to the swapping of variable, and \model stay above all the baseline for low values of $N$.

%% SARCOPENIA DATASET
For the sarcopenia dataset Figure~\ref{fig:sarcopenia:lower_supervision}, we can note that  \model performances decrease slower than the supervised baselines. We conjuncture that LSM is able to leverage the side information from only $x$ or only $y$ views without increasing the discrepancy between both domains. Where SOP learning schema does not take into account this discrepancy and falls behind the baseline.
We suppose that IODA is able to perform well in this kind of setting by providing a good parameter initialization without constraining the later learning of the model.

\begin{table}
\centering
\caption{mIoU evaluation on the \toy dataset according to the portion of available pairs $(x,y)$ from the dataset}
\label{table:exp:lower_supervision:toy}
\includegraphics[width=\textwidth]{table_data/toyset.tex}
\end{table}

%\label{table:exp:lower_supervision:sarcopenia}
\begin{table}
\centering
\caption{mIoU evaluation on the sarcopenia dataset according to the portion of available pairs $(x,y)$ from the dataset}
\label{table:exp:lower_supervision:sarcopenia}
\includegraphics[width=\textwidth]{table_data/sarcopenia.tex}
\end{table}

\section{Conclusions}
We presented \model, a novel training framework allowing to learn domain translation with multiple supervision levels, either fully supervised, or with views from only some domains. \model learns for each domain a latent space and a mapping function between each of these latent spaces. Unlike previous approaches, \model does not make the shared latent assumption directly, but rather relaxes it by supposing the existence of a mapping function between latent spaces. And further constraints each mapped latent code to be consistent with the original latent code from one domain. This makes \model suitable for multi-modal domain translation tasks, especially when incomplete supervision is available.

The experiments confirmed that \model is not penalizing when a lot of supervision is available and is able to leverage additional information in order to improve the training of models in low data settings.
%

% Future Work
While \model allows learning the translation between multiple domains in a weakly-supervised setting, it does not leverage the availability of multiple related views to enhance the translation during the inference (\model leverages the availability of multiple related views during training only).
We plan to address this issue in future works which will combine information from multiple related views as input to improve the translation.
Another issue that arises from \model is the increasing model size as the number of domains grows, this makes such an approach heavy for a high number of domains.

\section{Acknowledgments}
This work was financially supported by the ANR Labcom Lisa ANR-20-LCV1-0009. We also thank our colleagues at CRIANN, who provided us with the computation resources necessary for our experiments.

\bibliographystyle{splncs04}
\bibliography{b}

\appendix
\newpage
\section{Dataset Details}\label{app:dataset_details}

\subsection{\TOY (\toyS) Dataset}
\label{app:dataset_details:toyset}
The \toy dataset consists of image pairs made by superposing a background texture and a ring from another texture \footnote{https://www.ux.uis.no/$\sim$tranden/brodatz.html} for the input domain and the associated circle segmentation for the output domain.
Four parameters control the texture generation: circle x position, circle y position, circle radii, and circle thickness. In order to complexify the task, for the segmentation mask, we swapped the x position with the circle thickness and the y position with the circle radii. This changes the task from a segmentation task to a more challenging image-to-image translation task.
We use the standard mean Intersection over Union (mIoU) as evaluation metric for the output prediction and generate a total of 500 images.
Figure~\ref{fig:toyset_dataset} displays two pairs of input image and ground truth.
%%%%%
The images are $128 \times 128$ pixels (we randomly take a square in the texture 17 and 77 of size $128 \times 128$). Values are normalized in $[0,1]$, 500 samples are generated. We train on 70\% of the data, keep 10\% for the validation and 20\% for the testing. We use the cross-entropy as the loss function for the segmentation-related tasks, mean-square-error for input reconstruction-related tasks. As segmentation metric, we use the mean intersection-over-union, computed on the three classes and not on the background.

\begin{figure}
\centering
\includegraphics[height=2cm]{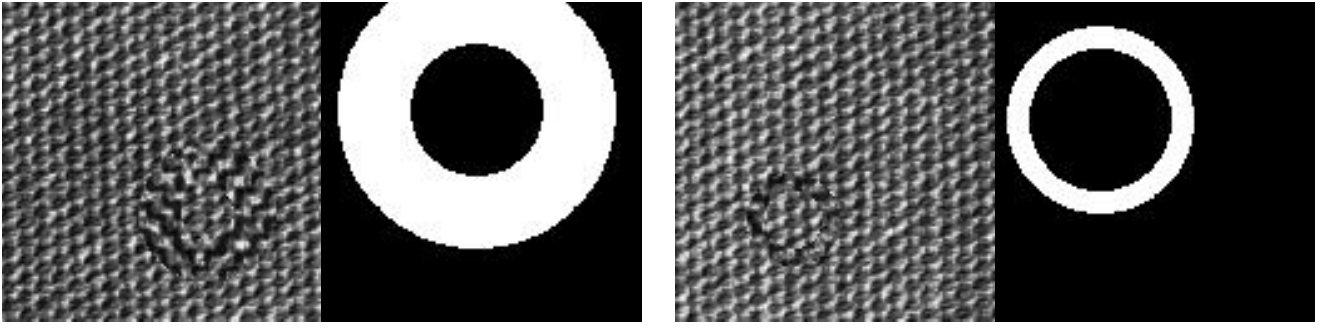}
\caption{Two couples of input image textures and their associated output. Here we swap generating parameter between the image and its segmentation creating an image-to-image translation task}
\label{fig:app:toyset_dataset}
\end{figure}

\subsection{Facial Landmark Detection (\faceS) Dataset}
\label{app:dataset_details:face}
We study the real case example of facial landmark detection, on the 300 Faces In-the-Wild Challenge (\faceS) \cite{Face_dataset_1,Face_dataset_2}. This 300 Faces In-the-Wild Challenge dataset is constituted of 300 indoor images and 300 outdoor images with associated 68 facial landmarks for each face. It provides a context with multiple modalities (image faces, and vector of positions) and with a highly structured output domain that presents strong intra-dependencies.

We rescale each image into $64 \times 64$ pixels and normalize the pixels values in $[0,1]$ and the landmarks in $[0,1]$. We train on 70\% of the data, keep 10\% for the validation, and 20\% for the testing. We use the mean-square error for input reconstruction-related tasks and prediction-related tasks. We use Normalized Root Mean Squared Error (NRMSE in equation~\ref{eq:NRMSE})\cite{NRMSE} as the metric for the prediction task.

\begin{align}
\label{eq:NRMSE}
    NRMSE(\hat y, y)=\frac{\sum^N_{i=1}\|\hat y_i-y_i\|_2}{N\times D}
\end{align}
With $N=68$ the number of landmark points $\hat y \in \mathbb{R}^{N\times 2}$ is the landmark prediction and $y \in \mathbb{R}^{N \times 2}$ is the landmark ground-truth, and $D$ is the inter-ocular distance computed from $y$.

\begin{figure}
\centering
\includegraphics[height=2.5cm]{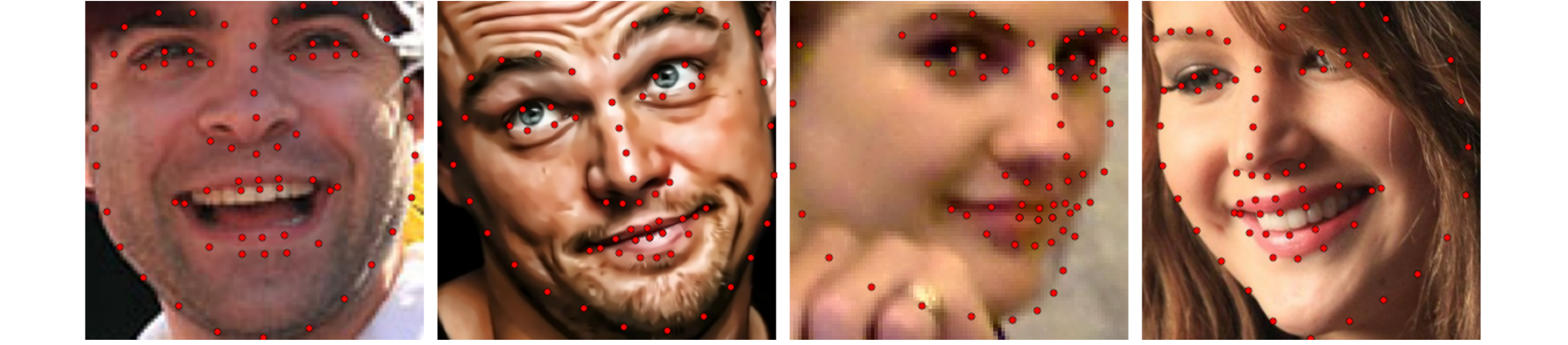}
\caption{Faces with their associated facial landmarks from the 300-W dataset}
\label{fig:app:face_dataset}
\end{figure}

\subsection{Sarcopenia Semantic Segmentation Dataset (\sarcoS)}
\label{app:dataset_details:sarcopenia}
Segmentation of the third lumbar vertebra (L3) level is useful in order to predict, from CT scan (computed tomography scan), the sarcopenia level - characterizing a level of muscle atrophy - which is later used in cancer diagnostic \cite{sarco_cancer_1,sarco_cancer_2}.
The segmentation task is a tedious task for the medical practitioner as the hand-made segmentation can take 4 minutes for an expert\cite{IODA}.
Two challenges arise from the data, inter-patient variability, as there are different factors that will impact the CT scan (age, sex, disease \dots); and inter-image variability, a lot of parameters exterior the patient can impact the CT scan quality (quality of the device, the radiation dosage \dots). This is illustrated in Figure~\ref{fig:sarcopenia_dataset} with two samples of CT scans and their ground-truth segmentation that are really different from one another.
Examples of data are given in Figure~\ref{fig:sarcopenia_dataset}, there are 4 classes: background (black), subcutaneous adipose tissue (purple), visceral adipose tissue (green), and skeletal muscle (red). The skeletal muscle segmentation is used to compute the sarcopenia level.
The dataset contains a total of 527 pairs of CT images and their masks, we train on 70\% of the data, keeping 10\% for the validation and 20\% for the testing. For this task, the image's background is cropped, they are resized to $128 \times 128$ pixels and normalized between 0 and 1. We use the cross-entropy as the loss function for the segmentation-related tasks, mean-square-error for input reconstruction-related tasks. As segmentation metric, we use the mean intersection-over-union, computed on the three classes and not on the background.

\begin{figure}
\centering
\includegraphics[height=2.5cm]{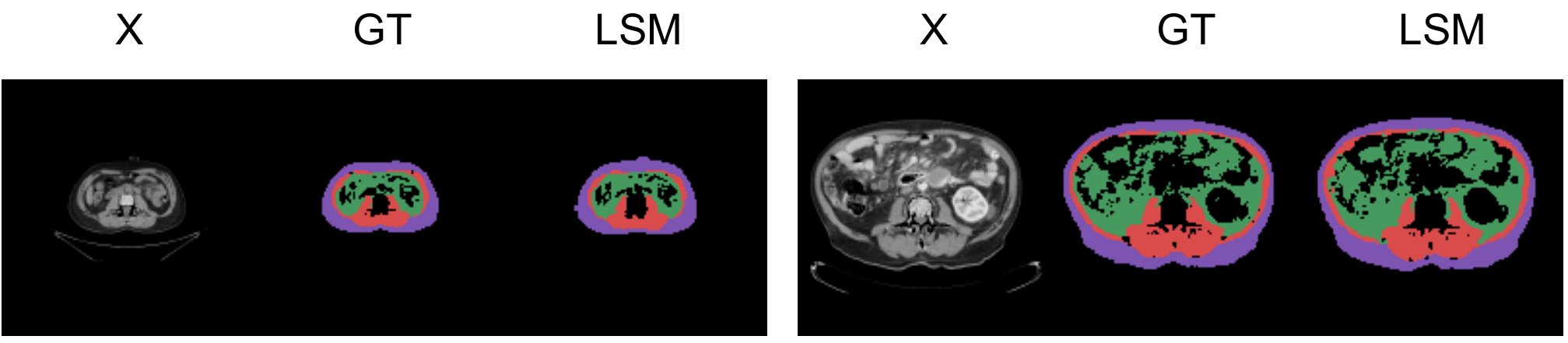}
\caption{Two tuples of CT scans, their ground-truth segmentation mask from our dataset and the \model translation}
\label{fig:app:sarcopenia_dataset}
\end{figure}

\section{Experimental Settings Details} \label{app:exp_det}

%\subsection{Architecture Details}
%\label{app:architecture_details}
We reuse code and architecture from CycleGAN and Pix2Pix implementation\footnote{\url{https://github.com/junyanz/pytorch-CycleGAN-and-pix2pix}}. The encoders-decoders are made of resnet 6-blocks\cite{cyclegan}, and the discriminators are made of PatchGAN~\cite{patchgan} for the \sarcoS and \toyS datasets.

For the \faceS dataset and the facial landmark domain (made of 68 2d coordinates), encoders are multilayer perceptron of dimension $[68 \times 2,64, 256, 1024]$ and the opposite for the decoders. The link network $\Lxy$ is made of a flattening layer, followed by a linear layer.

For Pix2Pix, for the \sarcoS and \toyS datasets, the output and the input are concatenated before going through the discriminator. For the \faceS dataset where the outputs are facial landmarks, they are repeated and concatenated on each spatial dimension of the input image before going through the discriminator.

We use Adam optimizer with $\beta_1=0.9$ and $\beta_2=0.999$ for all datasets and set the learning rate as a hyperparameter to optimize. For the \sarcoS dataset, \toyS dataset, and \faceS dataset respectively batch sizes of 16, 16, and 32 are used. We allow a computational budget of 1000 epochs with early stopping and patience of 100 epochs. Hyperparameter tuning is done using bayesian optimization, for the learning rate within the range $[1e-6,1e-1]$ in log space, and for each lambda in the range $[1e-4,10]$ in log space. The search is then run for the proportions $N=100\%$, $50\%$, and $5\%$ on 40 runs each, we then use the best hyperparameters for each proportion of data.
Table~\ref{table:losses_choice} list the functions used for each loss and dataset. For further detail, please refer to the GitHub implementation\footnote{\code}.

\setlength{\tabcolsep}{4pt}
\begin{table}
\begin{center}
\caption{Used losses function for each dataset and branch}
\label{table:losses_choice}
\begin{tabular}{lllllll}
\hline\noalign{\smallskip}
Dataset & $\mathcal L_{x\rightarrow y}$ & $\mathcal L_{x\rightarrow x}$ & $\mathcal L_{y\rightarrow y}$ & $\lossdist$ & $\lossconf$ & $\mathcal L_{Dc}$\\
\noalign{\smallskip}
\hline
\noalign{\smallskip}
Sarcopenia             & CE & MSE & CE & $L_1$ & BCE vs Uniform & BCE\\
Synthetic Segmentation & BCE & MSE & BCE & $L_1$ & BCE vs Uniform & BCE\\
Facial Landmark        & MSE & MSE & MSE & $L_1$ & BCE vs Uniform & BCE\\

\hline
\end{tabular}
\end{center}
\end{table}
\setlength{\tabcolsep}{1.4pt}

\section{\model Training complexity}
\begin{figure}[H]
\centering
\includegraphics[height=6cm]{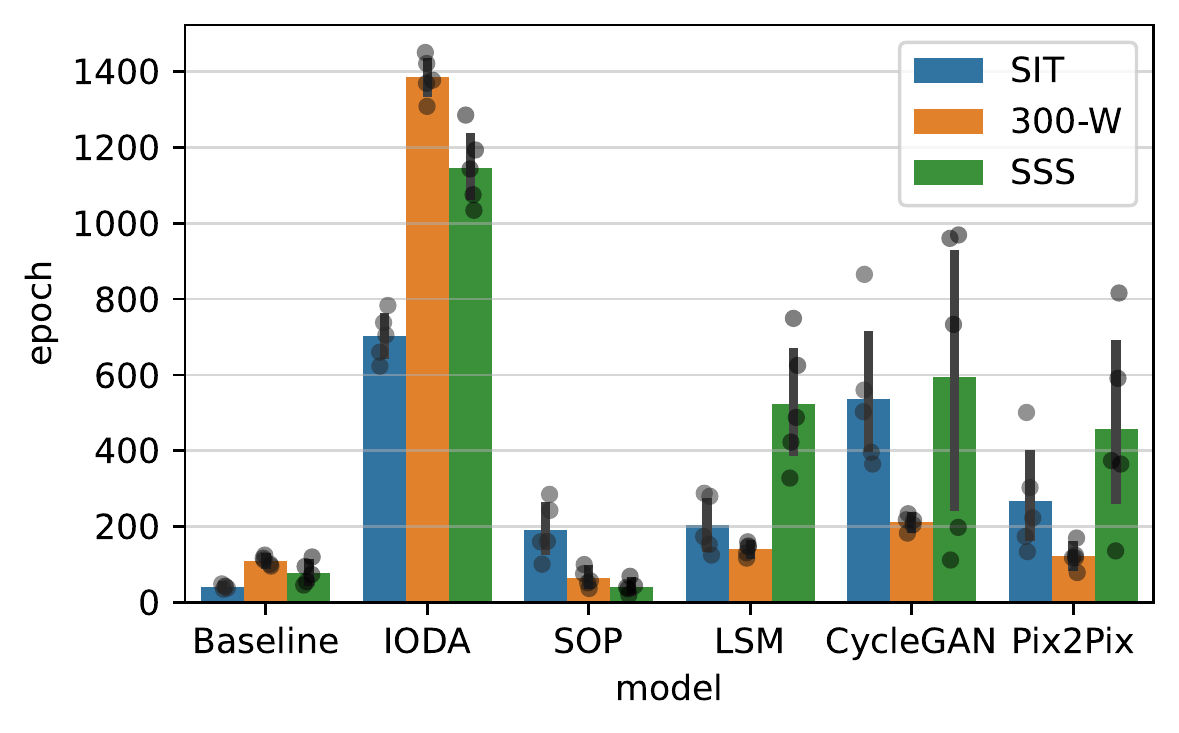}
\caption{Indices of the best epoch per model and per dataset using the fully labeled dataset ($N=100$).}
\label{fig:epochs:all}
\end{figure}

We study the possible training overhead that \model might induce in terms of the number of epochs. For each model, a training budget of 1000 epochs was available and for IODA, we also allow 1000 epochs for each pre-training stage. We compare how the different approaches behave within this computational budget and for each dataset. We display results in Figure~\ref{fig:epochs:all}.

 We did not find that \model uses an excessive number of epochs before early stopping (with a patience of 100 epochs), especially when compared to pre-training methods such as IODA, with an additional training stage, which can take a lot of training.
 
For methods using Generative Adversarial Networks such as CycleGAN and Pix2Pix, we consume all the training budget but display the best epoch per run. For unsupervised method as CycleGAN, it is not possible to perform early stopping as there is no ground truth available to compute a metric on an unpaired dataset in practice.

\section{Distance and Adversarial Loss Ablation}
We study how the different losses contribute to the model performances while varying the amount of available supervision. We name each ablation according to the format $\lambda_1\;\lambda_2\;\lambda_3\;\lambda_4$ where $\lambda_i=1$ if $\lambda_i \ne 0$. We found that, among all configurations, the one using every loss, has the lower NRMSE and the higher mIoU.

%\label{fig:ablation:losses:face}
\begin{figure}[H]
\centering
\includegraphics[height=4cm]{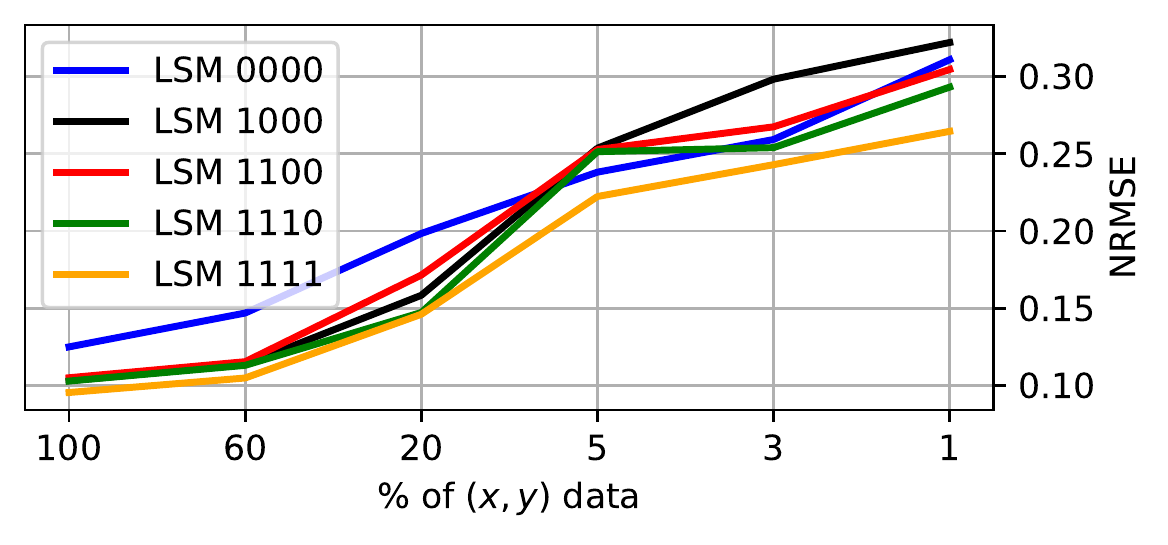}
\caption{Impact of the losses on \model on \faceS dataset on the NRMSE metric}
\label{fig:ablation:losses:face}
\end{figure}

%\label{table:exp:ablation:losses:face}
\setlength{\tabcolsep}{4pt}
\begin{table}
\begin{center}
\caption{Impact of losses ablation on \model, on \faceS dataset on the NRMSE metric}
\label{table:exp:ablation:losses:face}
%%%
\begin{tabular}{ lllllll }

\hline\noalign{\smallskip}
Model & 100 \% & 60 \% & 20 \% & 5 \% & 3 \% & 1 \% \\ \hline

\model 0000 & 1.25e-1 & 1.47e-1 & 1.98e-1 & \textit{2.38e-1} & 2.59e-1 & 3.11e-1 \\ 
\model 1000 & 1.03e-1 & 1.13e-1 & 1.58e-1 & 2.54e-1 & 2.98e-1 & 3.22e-1 \\ 
\model 1100 & 1.05e-1 & 1.15e-1 & 1.71e-1 & 2.53e-1 & 2.67e-1 & 3.05e-1 \\ 
\model 1110 & \textit{1.03e-1} & \textit{1.13e-1} & \textit{1.47e-1} & 2.51e-1 & \textit{2.54e-1} & \textit{2.93e-1} \\ 
\model 1111 & \textbf{9.53e-2} & \textbf{1.05e-1} & \textbf{1.46e-1} & \textbf{2.22e-1} & \textbf{2.43e-1} & \textbf{2.65e-1} \\ 
%%%

\noalign{\smallskip}
\hline
\noalign{\smallskip}
\end{tabular}
\end{center}
\end{table}
\setlength{\tabcolsep}{1.4pt}
%%%%%%%%%%%%%%%%%%%%%%%%%%%%%%%%%%%%%%%%%%%%%%%%%%%%%%%%%%%%
%\label{fig:ablation:losses:sarcopenia}
\begin{figure}[H]
\centering
\includegraphics[height=4cm]{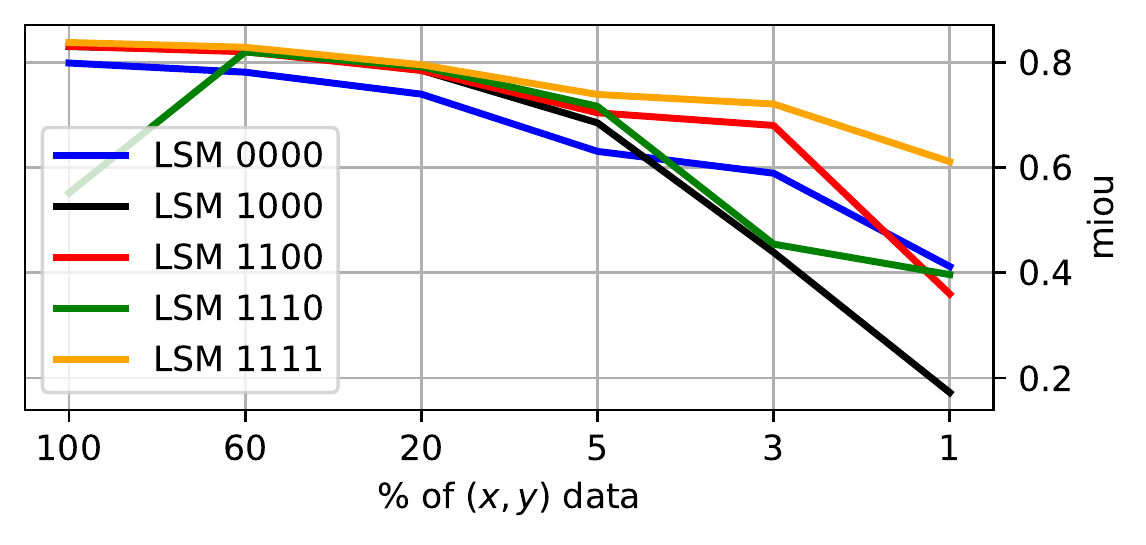}
\caption{Impact of the losses on \model on \sarcoS dataset on the mIoU metric}
\label{fig:ablation:losses:sarcopenia}
\end{figure}

%\label{table:exp:ablation:losses:sarcopenia}
\setlength{\tabcolsep}{4pt}
\begin{table}
\begin{center}
\caption{Impact of losses ablation on \model, on \sarcoS dataset on the mIoU metric}
\label{table:exp:ablation:losses:sarcopenia}
\begin{tabular}{ lllllll }

\hline\noalign{\smallskip}
Model & 100 \% & 60 \% & 20 \% & 5 \% & 3 \% & 1 \% \\ \hline

\model 0000 & 7.99e-1 & 7.81e-1 & 7.40e-1 & 6.31e-1 & 5.89e-1 & \textit{4.12e-1} \\ 
\model 1000 & 8.30e-1 & \textit{8.23e-1} & 7.85e-1 & 6.85e-1 & 4.39e-1 & 1.72e-1 \\ 
\model 1100 & \textit{8.31e-1} & 8.20e-1 & 7.85e-1 & 7.04e-1 & \textit{6.80e-1} & 3.60e-1 \\ 
\model 1110 & 5.52e-1 & 8.20e-1 & \textit{7.92e-1} & \textit{7.16e-1} & 4.54e-1 & 3.96e-1 \\ 
\model 1111 & \textbf{8.38e-1} & \textbf{8.29e-1} & \textbf{7.96e-1} & \textbf{7.39e-1} & \textbf{7.21e-1} & \textbf{6.11e-1} \\  
\noalign{\smallskip}
\hline
\noalign{\smallskip}
\end{tabular}
\end{center}
\end{table}
\setlength{\tabcolsep}{1.4pt}

%%%%%%%%%%%%%%%%%%%%%%%%%%%%%%%%%%%%%%%%%%%%%%%%%%%%%%%%%%%%

%\label{fig:ablation:losses:toy}
\begin{figure}[H]
\centering
\includegraphics[height=4cm]{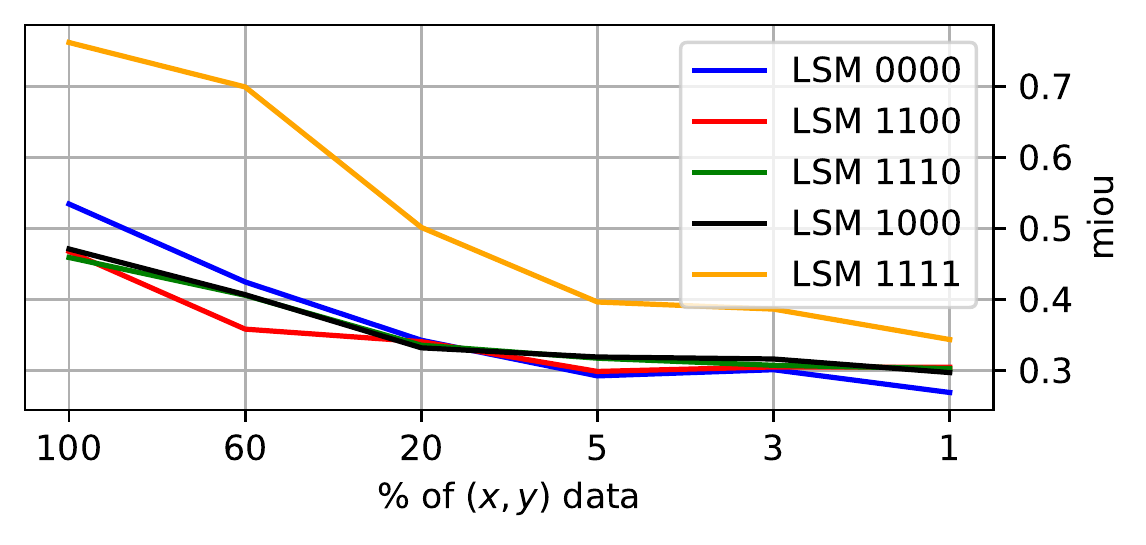}
\caption{Impact of the losses on \model on \toyS dataset on the mIoU metric}
\label{fig:ablation:losses:toy}
\end{figure}

%\label{table:exp:ablation:losses:toy}
\setlength{\tabcolsep}{4pt}
\begin{table}
\begin{center}
\caption{Impact of losses ablation on \model, on \toyS dataset on the mIoU metric}
\label{table:exp:ablation:losses:toy}
\begin{tabular}{ lllllll }

\hline\noalign{\smallskip}
Model & 100 \% & 60 \% & 20 \% & 5 \% & 3 \% & 1 \% \\ \hline

\model 0000 & \textit{5.35e-1} & \textit{4.25e-1} & \textit{3.43e-1} & 2.93e-1 & 3.01e-1 & 2.69e-1 \\ 
\model 1000 & 4.08e-1 & 3.76e-1 & 3.26e-1 & 3.16e-1 & \textit{3.04e-1} & \textit{2.91e-1} \\ 
\model 1100 & 3.99e-1 & 3.56e-1 & 3.26e-1 & 2.96e-1 & 3.00e-1 & 2.88e-1 \\ 
\model 1110 & 4.03e-1 & 3.73e-1 & 3.23e-1 & \textit{3.16e-1} & 3.00e-1 & 2.90e-1 \\ 
\model 1111 & \textbf{7.32e-1} & \textbf{6.50e-1} & \textbf{4.94e-1} & \textbf{3.85e-1} & \textbf{3.75e-1} & \textbf{3.27e-1} \\ 
\noalign{\smallskip}
\hline
\noalign{\smallskip}
\end{tabular}
\end{center}
\end{table}
\setlength{\tabcolsep}{1.4pt}

\section{Qualitative Results}
\label{app:qualitative}

We provide \model prediction for each dataset and different levels of supervision to access the evolution of prediction quality according to the available supervision.

\begin{figure}[H]
\centering
\includegraphics[height=13cm]{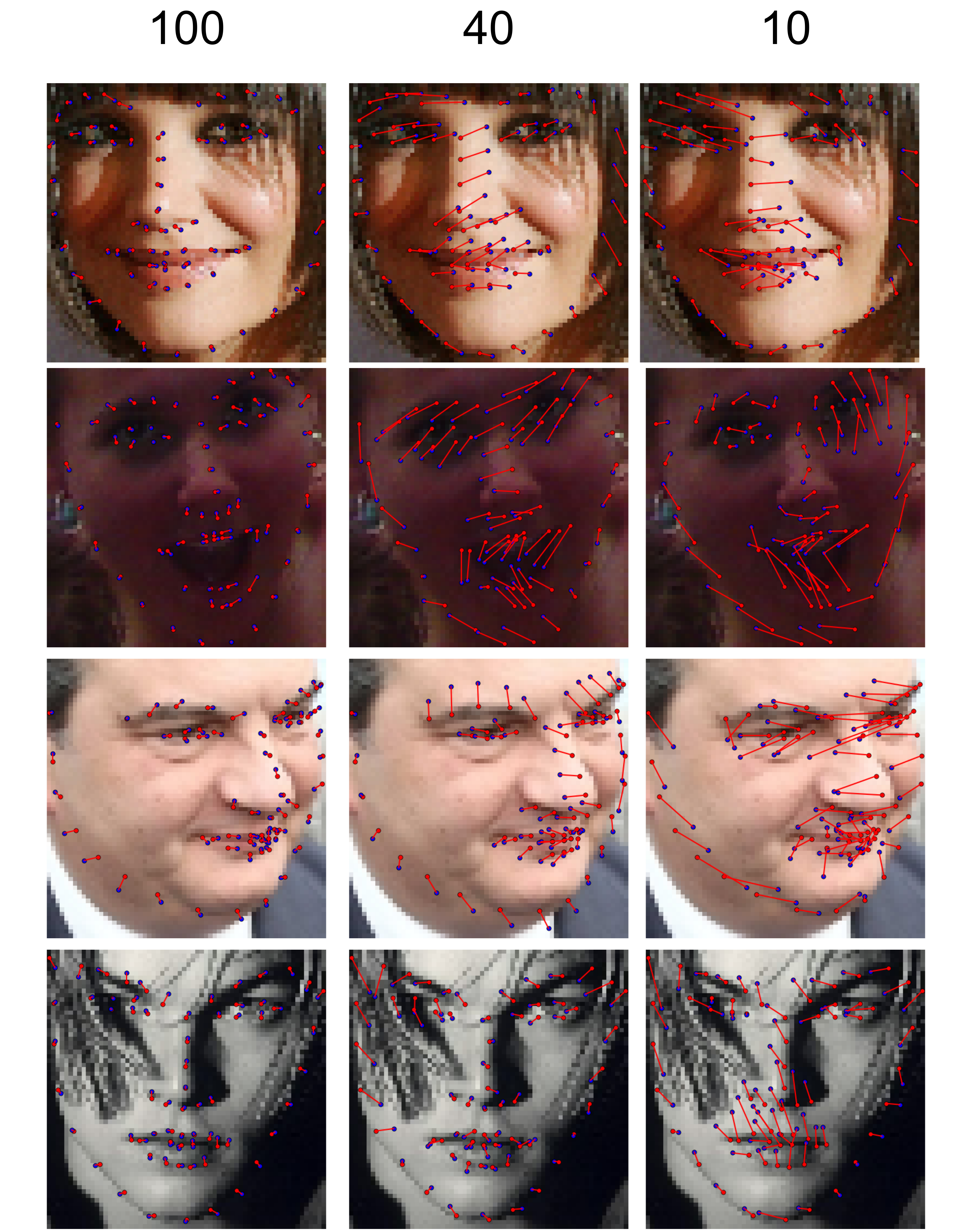}
\caption{Predicted sample from \model on \faceS test set, with different proportion of $(x,y)$ available (100\%, 40\%, 10\%). Ground-truth marker in blue, prediction marker in red}
\label{fig:app:face:pred}
\end{figure}
%%%%%%%%%%%%%%%%%%%%%%%%%%%%%%%%%%%%%%%%%%%%%%%%%%%%%%%%%
\begin{figure}[H]
\centering
\includegraphics[height=10cm]{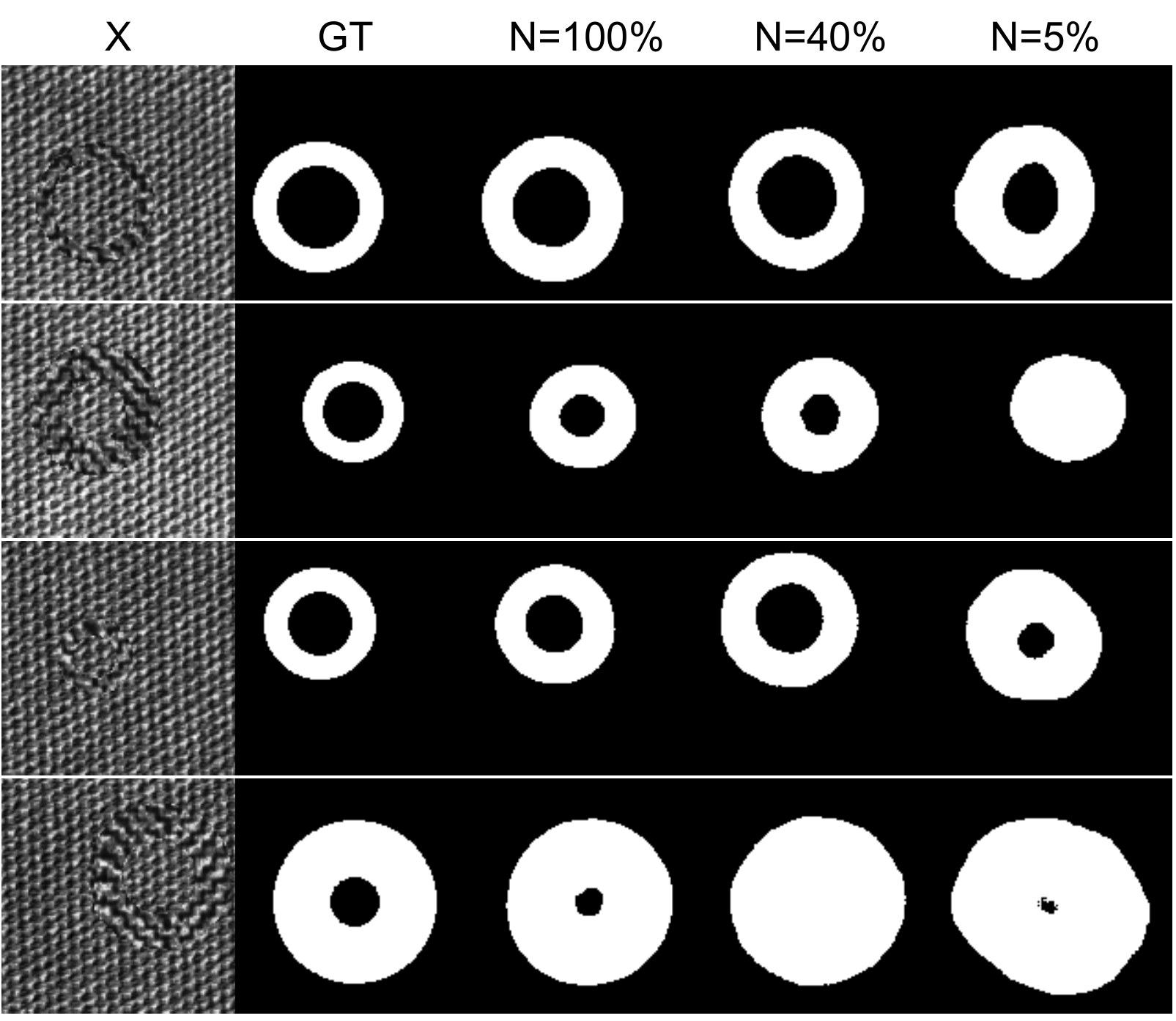}
\caption{Predicted sample from \model on \toyS test set, with different proportion of $(x,y)$ available (100\%, 40\%, 5\%)}
\label{fig:toy:pred_04}
\end{figure}

%%%%%%%%%%%%%%%%%%%%%%%%%%%%%%%%%
\begin{figure}[H]
\centering
\includegraphics[height=10cm]{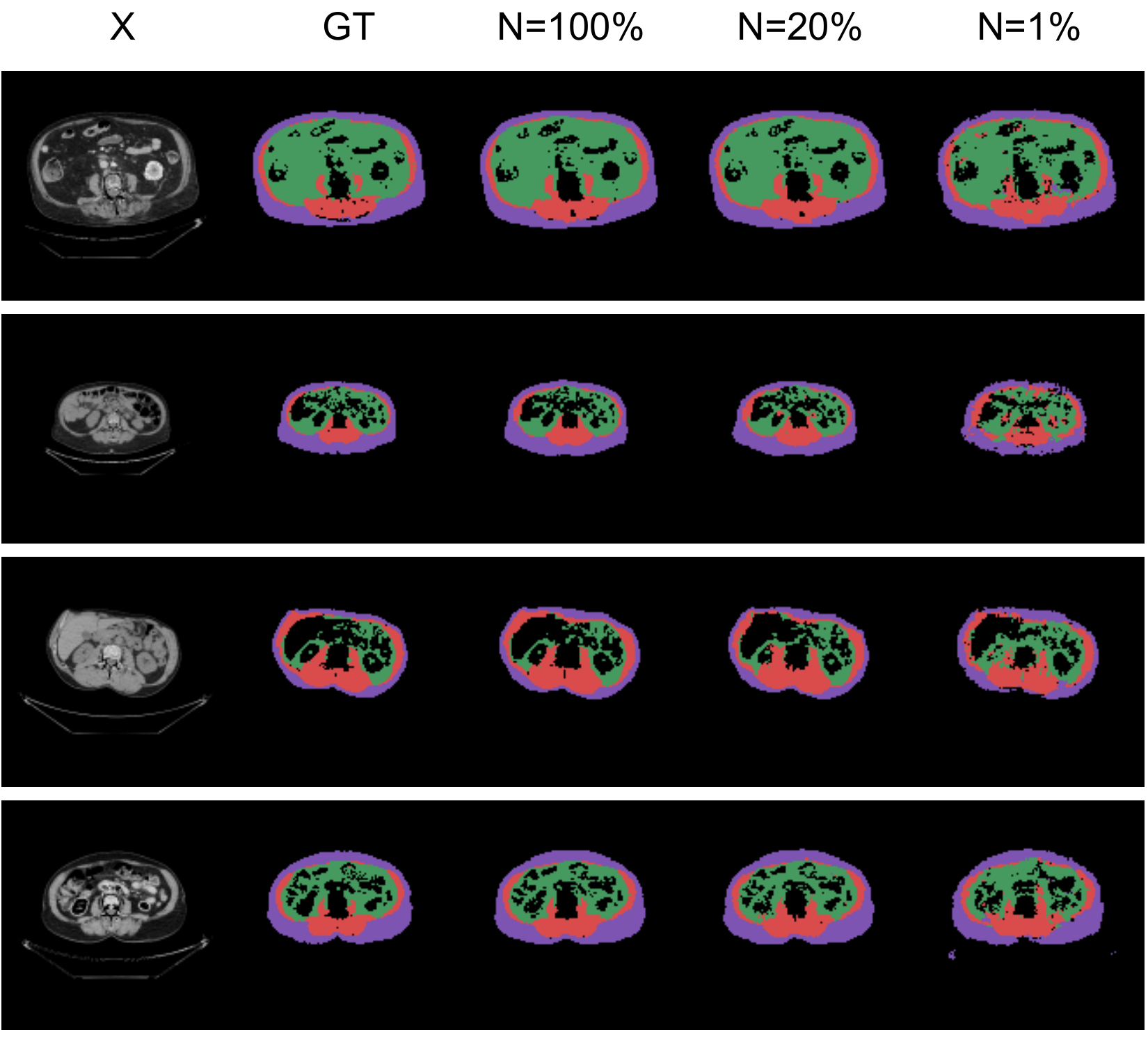}
\caption{Predicted sample from \model on \sarcoS test set, with different proportion of $(x,y)$ available (100\%, 20\%, 1\%)}
\label{fig:sarco:pred_full}
\end{figure}

\begin{figure}[H]
\centering
\includegraphics[height=12cm]{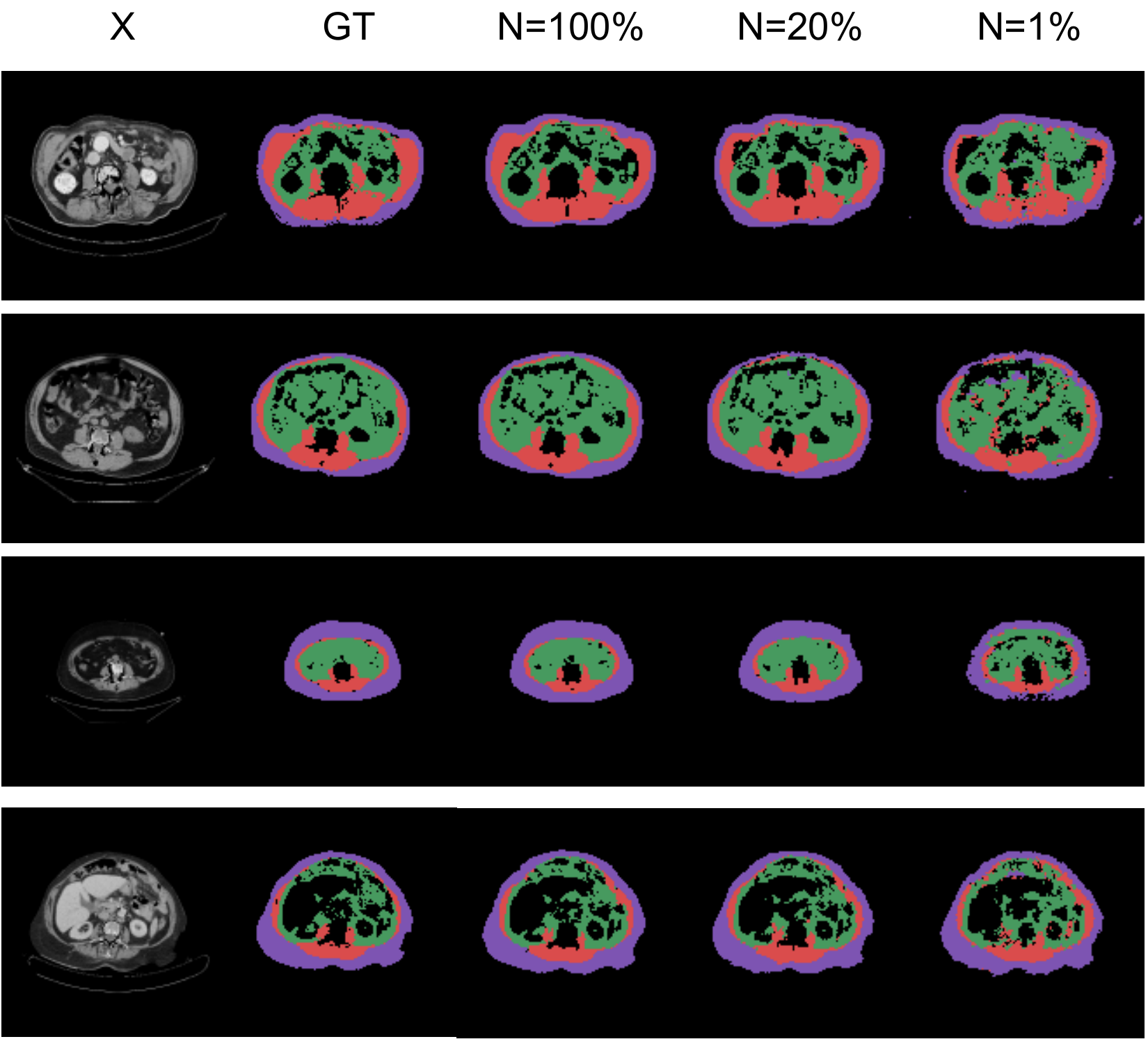}
\caption{Predicted sample from \model on \sarcoS test set, with different proportion of $(x,y)$ available (100\%, 20\%, 1\%)}
\label{fig:sarco:pred_1}
\end{figure}

\begin{figure}[H]
\centering
\includegraphics[height=12cm]{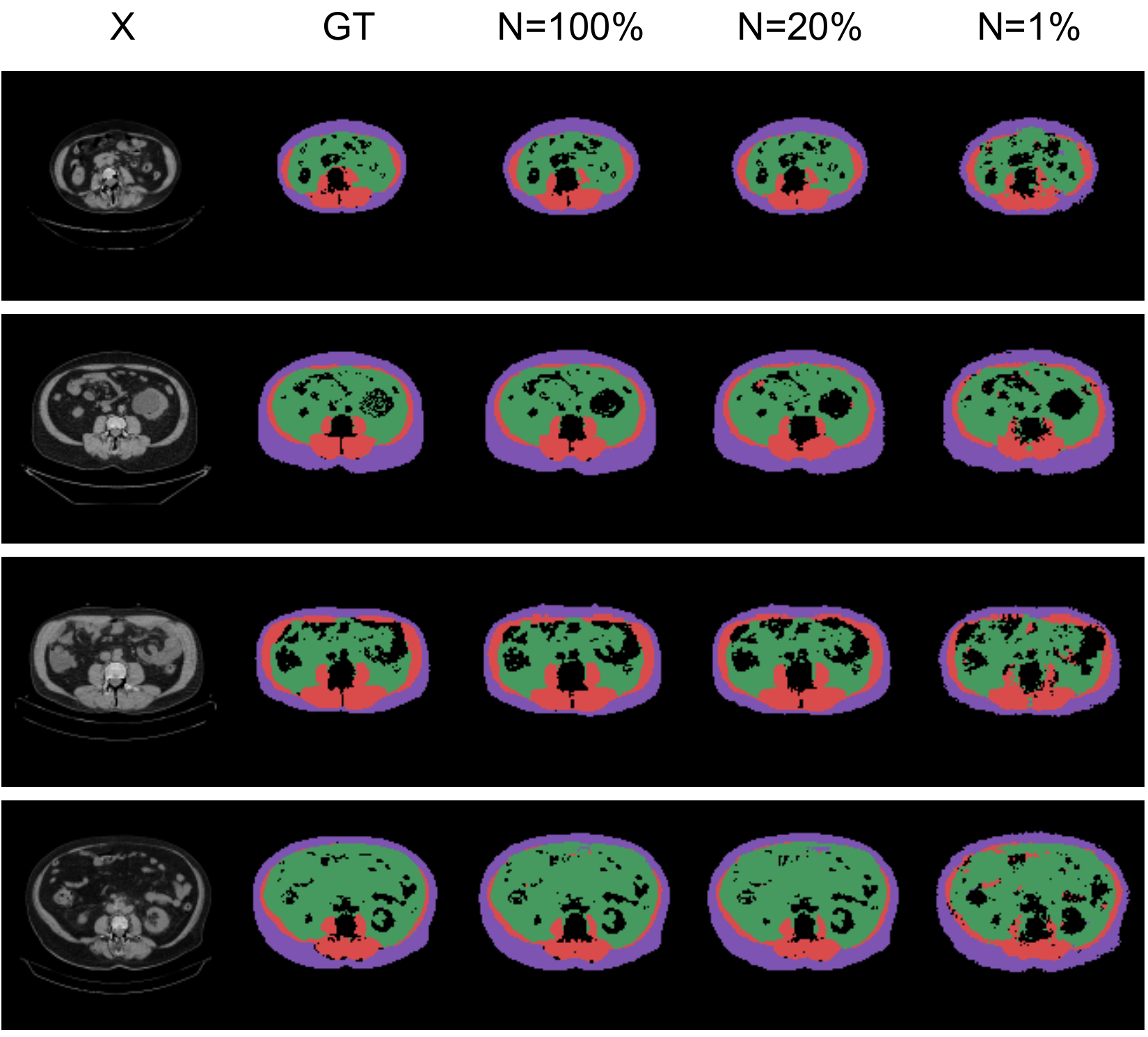}
\caption{Predicted sample from \model on \sarcoS test set, with different proportion of $(x,y)$ available (100\%, 20\%, 1\%)}
\label{fig:sarco:pred_02}
\end{figure}

\begin{figure}[H]
\centering
\includegraphics[height=12cm]{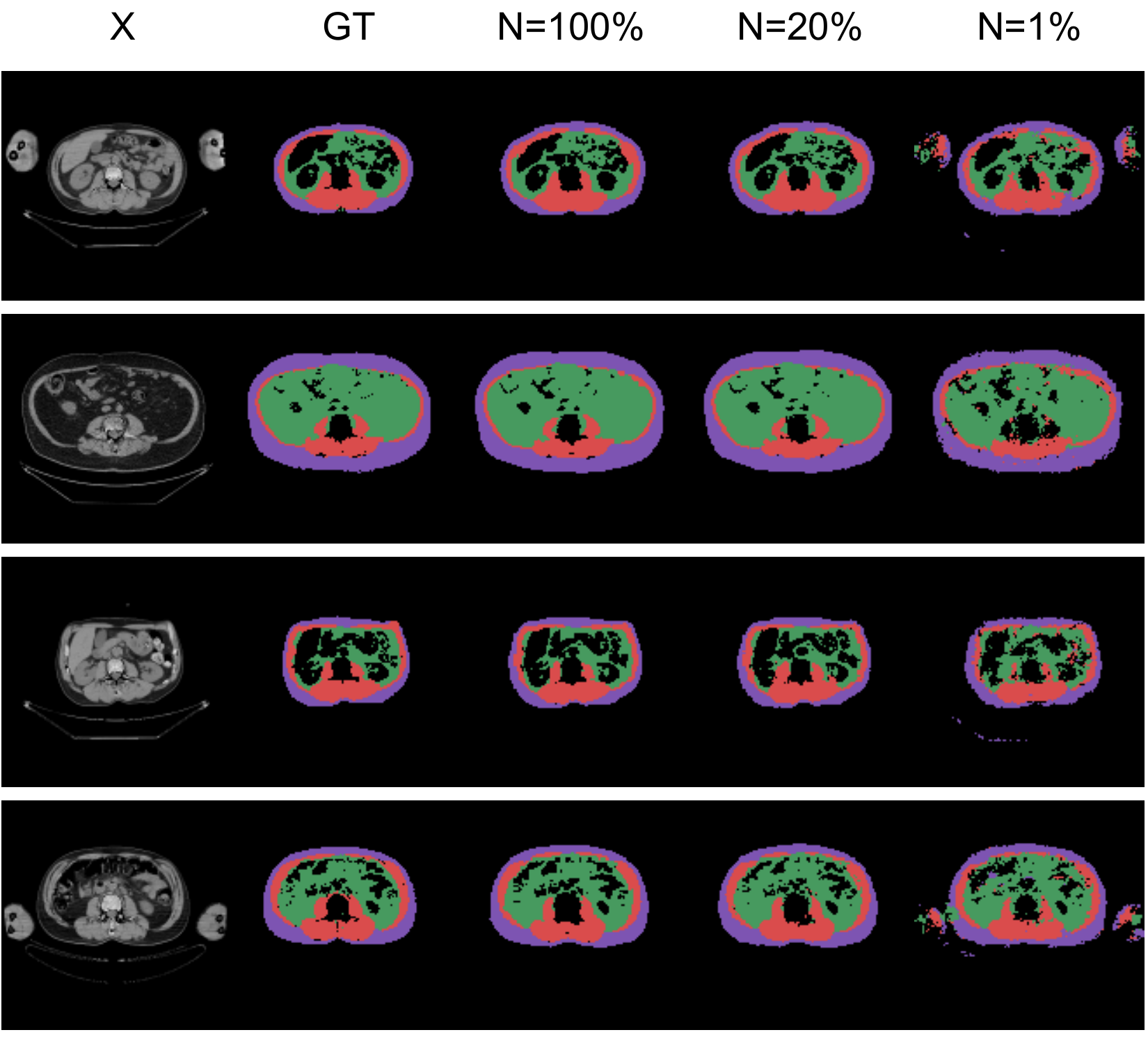}
\caption{Predicted sample from \model on \sarcoS test set, with different proportion of $(x,y)$ available (100\%, 20\%, 1\%)}
\label{fig:sarco:pred_005}
\end{figure}

\end{document}